\documentclass[runningheads]{llncs}

\usepackage{eccv}

\usepackage{eccvabbrv}

\usepackage{graphicx}
\usepackage{booktabs}

\usepackage[accsupp]{axessibility}  

\usepackage{hyperref}
\definecolor{cvprblue}{rgb}{0.21,0.49,0.74}
\definecolor{c1}{RGB}{139,127,192}
\definecolor{c2}{RGB}{66,133,244}
\definecolor{c3}{RGB}{220,135,67}
\usepackage{amsmath}
\usepackage{amssymb}
\usepackage{xcolor}
\usepackage{indentfirst}
\usepackage{caption}
\usepackage{multirow}
\usepackage{booktabs}
\usepackage{algorithm}
\usepackage{algorithmic}
\usepackage{graphicx}
\usepackage{marvosym}
\usepackage{orcidlink}

\begin{document}

\title{Diffusion Models as Optimizers for Efficient Planning in Offline RL} 


\author{Renming Huang\inst{1} \and
Yunqiang Pei\inst{1} \and
Guoqing Wang\inst{1}\textsuperscript{(\Letter)} \and \\
Yangming Zhang\inst{1} \and
Yang Yang\inst{1} \and
Peng Wang\inst{1} \and
Hengtao Shen\inst{2,1}}

\authorrunning{R. Huang et al.}

\institute{University of Electronic Science and Technology of China\\
\email{hrenming13@gmail.com, peiyunqiang\_cam@163.com,\\
p.wang6@hotmail.com, \{gqwang0420, yang.yang\}@uestc.edu.cn, } \and
Tongji University \\
\email{shenhengtao@hotmail.com}
}

\maketitle

\begin{abstract}
Diffusion models have shown strong competitiveness in offline reinforcement learning tasks by formulating decision-making as sequential generation.  However, the practicality of these methods is limited due to the lengthy inference processes they require.  In this paper, we address this problem by decomposing the sampling process of diffusion models into two decoupled subprocesses: 1) generating a feasible trajectory, which is a time-consuming process, and 2) optimizing the trajectory. With this decomposition approach, we are able to partially separate efficiency and quality factors, enabling us to simultaneously gain efficiency advantages and ensure quality assurance. We propose the Trajectory Diffuser, which utilizes a faster autoregressive model to handle the generation of feasible trajectories while retaining the trajectory optimization process of diffusion models.  This allows us to achieve more efficient planning without sacrificing capability.  To evaluate the effectiveness and efficiency of the Trajectory Diffuser, we conduct experiments on the D4RL benchmarks.  The results demonstrate that our method achieves $\it 3$-$\it 10 \times$ faster inference speed compared to previous sequence modeling methods, while also outperforming them in terms of overall performance. \url{https://github.com/RenMing-Huang/TrajectoryDiffuser}
\end{abstract}

\section{Introduction}
\label{sec:intro}
Offline Reinforcement Learning (Offline RL) is a data-driven RL paradigm concerned with learning exclusively from static datasets of previously collected experiences~\cite{levine2020offline}. In this setting, a behavior policy interacts with the environment to collect a set of experiences, which later can be used to learn a policy without further interaction. This paradigm is extremely valuable in settings where online interaction is impractical, either because data collection is expensive or uncontrollable. Learning optimal policies within a limited dataset presents significant challenges, the policy needs to effectively handle out-of-distribution (OOD) data and possess the capability to generalize to novel environments that are not encountered within the offline dataset~\cite{levine2020offline}.
\begin{figure}[t]
    \includegraphics[width=\textwidth]{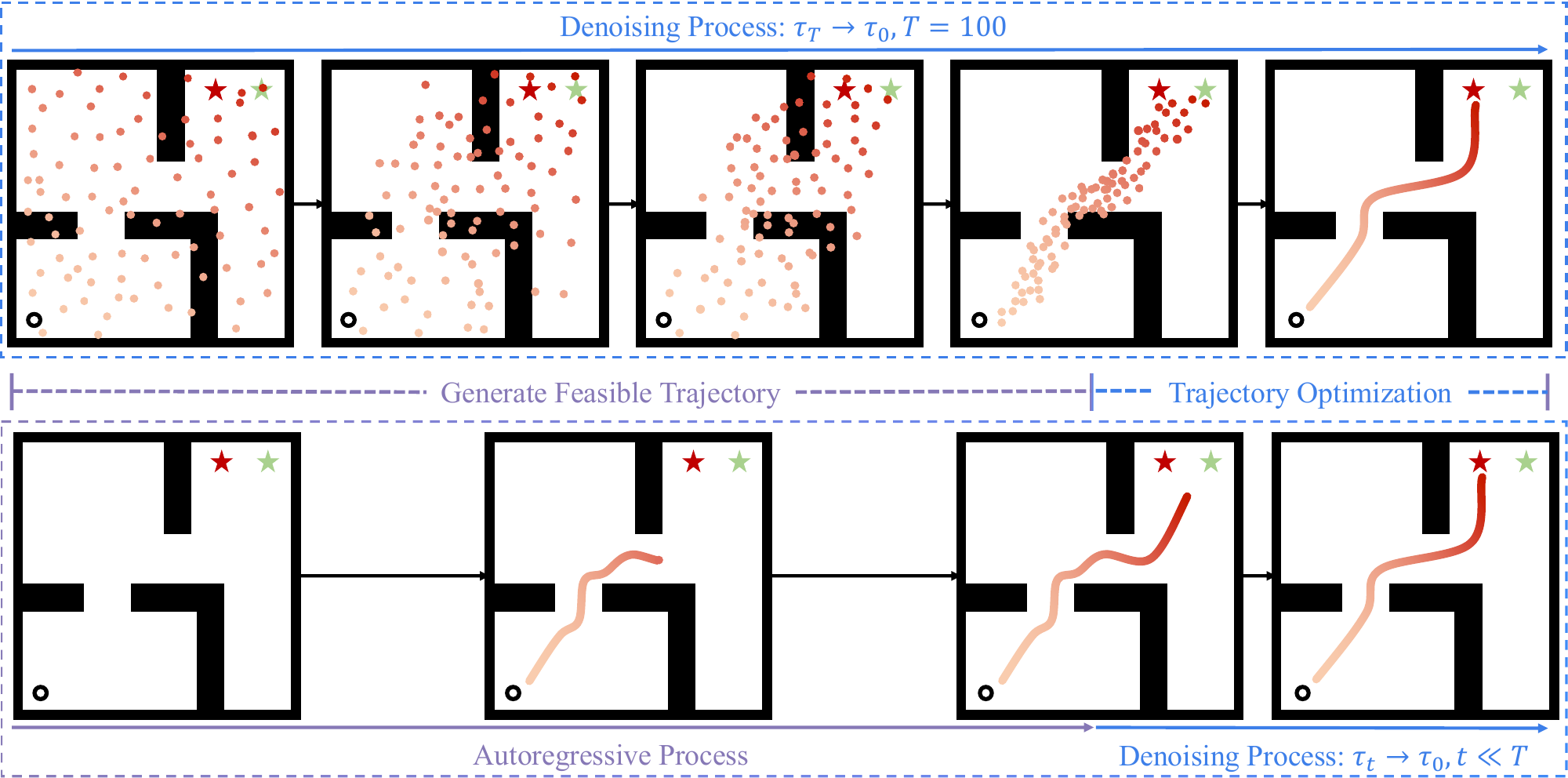}
    \captionof{figure}{\textbf{Illustrative Example and Motivation.} We  visualize the trajectory planning process of a maze task, where the agent needs to navigate to the higher-reward \textcolor{red!60!black}{red} target. In the first row, we elucidate that the denoising process can be decomposed into two components: \textcolor{c1}{Generate Feasible Trajectory} and \textcolor{c2}{Trajectory Optimization}. In the second row, we illustrate how an autoregressive process is employed to expedite the ``Generate Feasible Trajectory" phase.}
    \label{example}
\end{figure}
Over the last few years, diffusion models~\cite{saharia2022photorealistic,nichol2021glide,nichol2021improved} have achieved remarkable success in the field of conditional generative modeling~\cite{saharia2022photorealistic, ramesh2022hierarchical}, showcasing their strong generalization abilities. This sparks significant interest in applying diffusion models to decision-making research. Previous studies~\cite{janner2022planning,ajay2022conditional} involve training generative trajectory models on offline datasets and employing them to generate trajectories based on conditions, which guide the decision-making process. Furthermore, some methods~\cite{liang2023adaptdiffuser} focus on improving diffusion models by generating synthetic data and utilizing reward signals to enhance the performance and adaptability on both known and unknown tasks.

The use of diffusion models for trajectory planning undoubtedly achieves significant success. However, these methods~\cite{janner2022planning,ajay2022conditional, liang2023adaptdiffuser,ni2023metadiffuser} suffer from the common challenge that limits their practicality: The diffusion models require lengthy denoising processes during inference, rendering them unsuitable for tasks with rapidly changing environments, such as autonomous driving. While there are attempts to address this problem in previous works~\cite{song2020denoising, nichol2021improved}, they often sacrifice the quality of sampling.

After an in-depth investigation into the denoising process, we have identified that the diffusion model employed for trajectory planning can be effectively decomposed into two distinct steps. First, there is the generation of a feasible trajectory, where the agent continuously optimizes the trajectory over time using the diffusion model. This initial optimization takes a substantial amount of computing time, gradually refining the trajectory into a reasonable configuration. Subsequently, in the last few steps of the diffusion process, the model explicitly optimizes the trajectory towards a higher-reward target, as illustrated in \Cref{example}.
This observation consolidates our understanding of how diffusion models operate in trajectory generation and motivates us to devise a more efficient approach to accelerate the denoising process. Our goal is to reduce the time required for the initial trajectory generation while offering a quality trajectory initialization to maintain or even improve the quality of the trajectory generated.

\begin{figure}[t]
    \centering
    \includegraphics[width=0.6\linewidth]{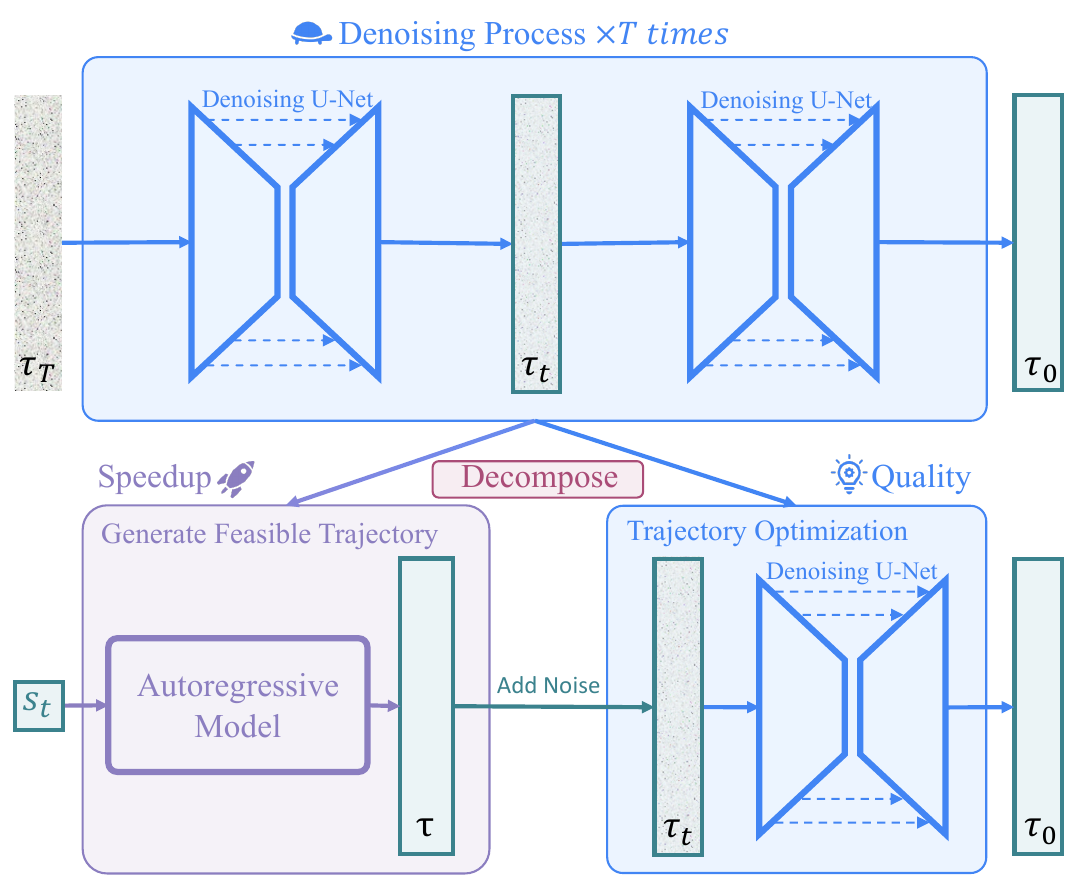}
    \caption{\textbf{Decompose the Trajectory Generation Process.} The denoising process is divided into two steps. To expedite the inference process, a more efficient autoregressive model is utilized to generate the feasible trajectory. A small portion of the denoising process is retained for trajectory optimization. This decomposition allows us to achieve improvements in both efficiency and performance simultaneously.}

    \label{teaser}
\end{figure}
Motivated by the above observation, we propose a method to accelerate trajectory generation without sacrificing sampling quality, named \textit{Trajectory Diffuser}, as illustrated in \Cref{teaser}. \textbf{To accelerate sampling}, we opt to employ a faster and simpler method to replace the first step of the process mentioned earlier. Drawing inspiration from the Decision Transformer~\cite{chen2021decision}, we utilize a transformer~\cite{vaswani2017attention} model with contextual awareness  generating a high-quality initial trajectory in an autoregressive manner and enjoying higher efficiency compared to the diffusion models.
\textbf{To guarantee trajectory quality}, despite the preservation of the diffusion model as the trajectory optimizer, only the final one-tenth of the denoising steps are leveraged as the optimization process. Additionally, we use conditional information to guide and improve the quality of the trajectory.

In summary, our contributions are three-fold: 

\begin{itemize}
\item  We decompose the inference process of the diffusion model into two steps: 1) generate a feasible trajectory, and 2) optimize the feasible trajectory. This decomposition allows us to separately consider the efficiency and quality of the generative model.

\item We introduce \textit{Trajectory Diffuser}, a novel offline reinforcement learning algorithm that not only mitigates the computational challenges of inference using diffusion models for trajectory planning but also delivers notable performance improvements.

\item We evaluate the efficiency and performance of Trajectory Diffuser on the D4RL benchmark tasks~\cite{fu2020d4rl} for offline RL. Compared to previous methods based on diffusion models, our approach achieves a $3$-$10\times$ improvement in efficiency while surpassing their performance.
\end{itemize}

\section{Related Work}
\label{sec:related work}
\subsection{Offline Reinforcement Learning}
Offline reinforcement learning is a technique that enables learning optimal policies from pre-collected datasets without requiring additional interaction. However, the distributional shift between the learned policy and data collection policy poses a significant challenge to improving performance~\cite{fujimoto2019off,wu2019behavior}. This disparity can cause overestimation of out-of-distribution behavior and inaccurate policy evaluations, leading to performance degradation~\cite{fujimoto2019off,kumar2020conservative}. Recent research in offline RL has identified two categories of methods to address this challenge: model-free and model-based approaches. Model-free techniques impose constraints on the learned policy and value function to prevent inaccurate estimation of unseen behavior and quantify the robustness of out-of-distribution behavior by introducing uncertainty~\cite{kostrikov2021offline,kumar2020conservative}. In contrast, model-based methods propose RL algorithms that learn the optimal policy through planning or synthetic experience generated from a learned dynamic model~\cite{hafner2019learning,nagabandi2018neural}. Despite a variety of techniques being proposed, the distributional shift challenge in offline RL remains an active research area, and the development of new algorithms and techniques is crucial for improving performance in practical environments.

\subsection{Diffusion Models in Reinforcement Learning}
Diffusion models have shown promising results in the learning of generative models for both image and text data~\cite{saharia2022photorealistic,nichol2021glide,nichol2021improved}. The iterative denoising process can be interpreted as optimizing the gradient of a parameterized data distribution to match a scoring objective~\cite{song2020denoising}, leading to the formation of an energy model~\cite{du2019implicit,nijkamp2019learning,grathwohl2020learning}. Diffusion-QL~\cite{wang2022diffusion} utilizes a diffusion model for policy regularization, effectively enhancing the expressive power of the policy by integrating guidance from Q-learning. Recent approaches extend diffusion models to trajectory generation and demonstrate their ability to model data distributions effectively~\cite{janner2022planning,ajay2022conditional}. Furthermore, AdaptDiffuser~\cite{liang2023adaptdiffuser} enhances the performance and adaptability of the diffusion model on known and unknown tasks by generating synthetic data and employing reward guidance. MetaDiffuser~\cite{ni2023metadiffuser} leverages the diffusion model as a conditional planner to achieve generalization across diverse tasks. However, these previous methods involved reverse diffusion starting from random noise, which result in high computational costs in terms of time. This limitation restricted their practicality.

\subsection{Transformers in Reinforcement Learning}
Transformers~\cite{vaswani2017attention} are a type of neural network architecture that has been successfully applied to various natural language processing~\cite{devlin2018bert,radford2018improving} and computer vision tasks~\cite{he2022masked,carion2020end}. However, their application in reinforcement learning has been relatively unexplored due to higher variance in training.~\cite{zambaldi2018deep,ritter2020rapid} have shown that augmenting transformers with relational reasoning and iterative self-attention can improve performance in combinatorial environments and enhance the utilization of episodic memories by RL agents. Moreover, new approaches that cast decision-making in RL as a sequence generation problem~\cite{zhang2023saformer,furuta2021generalized,chen2021decision,janner2022planning}, such as Decision Transformer (DT)~\cite{chen2021decision} and Trajectory Transformer (TT)~\cite{janner2021offline}, have shown great potential in improving RL performance. DT generates action based on historical state and reward token sequences, while TT discretizes such tokens and generates sequences using beam search. These developments indicate that transformers have the potential to be combined with RL and improve its performance.

\section{Preliminaries}
\label{sec:preliminaries}
\subsection{Offline RL}

In the field of sequential decision-making, the problem is formulated as a discounted Markov Decision Process defined by the tuple $\mathcal{M}= \langle \rho_0, \gamma, \mathcal{S, A, T, R} \rangle$, where $\rho_0$ represents the initial state distribution, $\mathcal{S}$ and $\mathcal{A}$ are the state and action spaces, $\mathcal{T : S \times A \rightarrow S}$ is the transition function, $\mathcal{R : S \times A \times S} \rightarrow \mathbb{R}$ gives the reward at any transition, and $\gamma \in [0, 1)$ is a discount factor~\cite{puterman2014markov}.The agent acts according to a stochastic policy $\pi :\mathcal{ S }\rightarrow \Delta_\mathcal{A}$, which generates a sequence of state-action-reward transitions, or trajectory $\tau = (s_k, a_k, r_k)_{k \geq 0}$, with a probability $p_\pi(\tau)$. The return $R(\tau) = \sum_{k \geq 0} \gamma^k r_k$ for a trajectory $\tau$ is defined as the sum of discounted rewards obtained along the trajectory. The standard objective in reinforcement learning is to find a policy $\pi^* = \arg \max_\pi \mathbb{E}_{\tau \sim p_\pi}[R(\tau)]$ that maximizes the expected return. This involves finding a return-maximizing policy from a fixed dataset of transitions collected by an unknown behavior policy $\mu$~\cite{levine2020offline}. Solving Offline RL using sequence modeling can be viewed as a trajectory optimization problem~\cite{janner2022planning}. The objective is to find an optimal sequence of actions $a^*_{0:T}$ that maximizes (or minimizes) an objective function $\mathcal{J}$, which is computed based on per-timestep rewards (or costs) $r(s_t, a_t)$:
\begin{equation}
    a_{0:T}^* = \arg\max_{a_{0:T}} \mathcal{J}(s_0, a_{0:T}) = \arg\max_{a_{0:T}} \sum_{t=0}^T r(s_t, a_t),
\end{equation}
the objective value of a trajectory is represented as $\mathcal{J}(\tau)$.

\subsection{Diffusion Probabilistic Models}
Diffusion Probabilistic models~\cite{sohl2015deep, ho2020denoising} are a type of generative model that learns the data distribution $q(x)$ from a dataset $\mathcal{D} := \{x_i\}_{0\leq i <M}$. It represents the process of generating data as an iterative denoising procedure, denoted by $p_\theta(x_{i-1}|x_i)$ where $i$ is an indicator of the diffusion timestep. The denoising process is the reverse of a forward diffusion process that corrupts input data by gradually adding noise and is typically denoted by $q(x_i|x_{i-1})$. The reverse process can be parameterized as Gaussian under the condition that the forward process obeys the normal distribution and the variance is small enough: $p_\theta(x_{i-1}|x_i) = \mathcal{N}(x_{i-1}|\mu_\theta(x_i,i),\Sigma_i)$, where $\mu_\theta$ and $\Sigma$ are the mean and covariance of the Gaussian distribution, respectively. The parameters $\theta$ of the diffusion model are optimized by minimizing the evidence lower bound of negative log-likelihood of $p_\theta(x_0)$, similar to the techniques used in variational Bayesian methods: $\theta^\ast = \arg\min_\theta -\mathbb{E}_{x_0}[\log p_\theta(x_0)]$. For model training, a simplified surrogate loss~\cite{ho2020denoising} is proposed based on the mean $\mu_\theta$ of $p_\theta(x_{i-1}|x_i)$, where the mean is predicted by minimizing the Euclidean distance between the target noise and the generated noise: $\mathcal{L}_{\text{denoise}}(\theta) = \mathbb{E}_{i,x_0 \sim q,\epsilon \sim \mathcal{N}}[|\epsilon - \epsilon_\theta(x_i,i)|^2]$, where $\epsilon\sim \mathcal{N}(0,\boldsymbol{I})$.

\section{Trajectory Diffuser}
\label{sec:method}
\begin{figure*}[t]
    \centering
    \includegraphics[width=\textwidth]{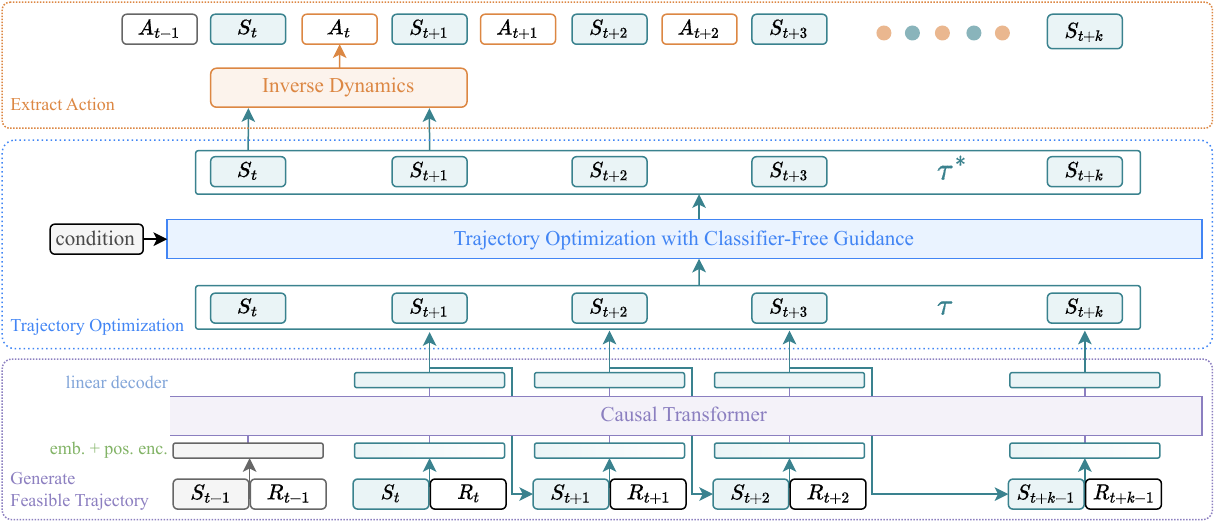}
    \caption{\textbf{Planning with Trajectory Diffuser.} Trajectory Diffuser first \textcolor{c1}{generates} the future trajectory $\tau$ autoregressively, denoted as $\tau=( S_{t+1}, \dots, S_{t+k})$, by considering the current state $S_t$ and the history states $( S_{t-h}, \dots, S_{t-1})$ combine with returns-to-go $(\hat{R}_{t-h}, \dots, \hat{R}_{t})$. Subsequently, it utilizes a diffusion model to \textcolor{c2}{optimize} $\tau \rightarrow \tau^*$, and employs inverse dynamics to \textcolor{c3}{extract} and execute the action $A_t$ that leads to the immediate future state $S_{t+1}$.} 
    \label{fig:overview}
\end{figure*}
In this section, we introduce Trajectory Diffuser, which reformulates the problem by decomposing the trajectory planning process of diffusion models into two steps: generate a feasible trajectory and trajectory optimization. Such design allows us to simultaneously ensure efficiency and performance while striking a balance, and we accordingly design a model consisting of two parts. In \Cref{feasible}, we employ a transformer-based autoregressive model to generate the feasible trajectory, thereby enhancing the inference speed. In \Cref{improve}, we retain the diffusion model for trajectory optimization, ensuring trajectory quality. The overview of our framework is summarized in the \Cref{fig:overview}.

\subsection{Trajectory Representation} 
The primary criterion for the selection of a trajectory representation is its capacity to facilitate meaningful pattern learning by the transformer model~\cite{chen2021decision,janner2021offline}. Decision Transformer~\cite{chen2021decision} provides us with a trajectory representation method with the capacity of learning to maximize returns-to-go $\hat{R}_t = \sum_{t'=t}^{T}r_{t'}$. In Decision Transformer, the last $K$ timesteps are fed into the Decision Transformer, for a total of $3K$ tokens (one for each modality: returns-to-go, state, or action). However, in the reinforcement learning setting, directly modeling actions suffers from some practical issues~\cite{janner2022planning}. First, while states are typically continuous in RL, actions are more varied and are often discrete. Furthermore, sequences over actions, which are often represented as joint torques, tend to be more high-frequency and less smooth, making them much harder to predict and model\cite{ajay2022conditional}. Therefore, our trajectory planning model excludes actions, resulting in the following trajectory representation $\tau$ being amenable to autoregressive training and generation:
\begin{equation}
   \tau = \left [ \begin{matrix}
    s_1& s_2& \dots &s_T\\
    \hat{R}_1 & \hat{R}_2 &\dots &\hat{R}_T \\
    \end{matrix} \right ] .
\end{equation}
\textbf{Acting with Inverse-Dynamics.} Merely sampling states from a diffusion model falls short of establishing a comprehensive controller. An inverse dynamics model~\cite{agrawal2016learning,pathak2018zero} takes two consecutive states as input and outputs the action that likely led to the transition from the first state to the second. Mathematically, this can be represented as follows:
Given two consecutive states $s_t$ and $s_{t+1}$, the inverse dynamics model $f_{\phi}$ with learnable parameters $\phi$ predicts the action $a_t$ that led to the transition from $s_t$ to $s_{t+1}$:
\begin{equation}
a_t = f_{\phi}(s_t, s_{t+1}).
\end{equation}

\subsection{Generate Feasible Trajectory with Transformer}\label{feasible}
Since trajectory has a natural ordering pattern, it is common to factorize the joint probabilities over symbols as the product of conditional probabilities. We train an autoregressive transformer model to predict the next state given the previous states and returns-to-go. The autoregressive transformer model takes as input a sequence of tokens of the form $(s_{t-h:t},\hat{R}_{t-h:t})$ and predicts the next state and expected reward $(s_{t+1},\hat{R}_{t+1})$, where $h$ stands for the history length. The model is trained to minimize the negative log-likelihood of the ground-truth sequence of states and rewards:
\begin{equation}
    \mathcal{L} = -\sum_{t=1}^T \log p(s_{t+1}, \hat{R}_{t+1}  | s_{t-h:t}, \hat{R}_{t-h:t}),
\end{equation}
where $p(s_{t+1},\hat{R}_{t+1} |s_{t-h:t},\hat{R}_{t-h:t})$ is the conditional probability distribution over the next state and expected reward given the previous states and rewards. 

To generate a trajectory using the trained Transformer model, we can start with history states $s_{t-h:t}$ and expected returns-to-go $\hat{R}_{t-h:t}$, We then repeatedly sample $l$ states and returns-to-go from the distribution generated by the model:
\begin{equation}
\begin{aligned}
(s_{t+1},\hat{R}_{t+1}) \sim p(s_{t+1},\hat{R}_{t+1} |s_{t-h:t},\hat{R}_{t-h:t}).
\end{aligned}
\end{equation}
\textbf{Architecture.} We employ the last $H$ timesteps, which comprise two types of information: state and returns-to-go. Unlike the decision transformer approach, we do not treat state and returns-to-go as separate tokens. Instead, we consider them as a unified entity, implicitly binding returns-to-go with state. In order to obtain token embeddings, we employ a training process that involves training distinct linear layers for each modality. These linear layers are responsible for converting the raw inputs into embeddings of the desired dimension. Subsequently, layer normalization is applied to ensure appropriate normalization of the embeddings. The processed tokens are then fed into a GPT model~\cite{radford2019language}, which utilizes autoregressive modeling to predict future states.\\
\textbf{Training.} We are provided with a dataset $\mathcal{D}$ of offline trajectories, we sample mini-batches of sequence length $K$ from the dataset. To prevent information leakage, we apply an upper triangular mask matrix to mask out all information beyond the current token. The prediction head associated with the input token $[s_t, \hat{R}_t]$ is trained to predict the next token, which consists of two components: the next timestep's state $s_{t+1}$ and reward-to-go $\hat{R}_{t+1}$. We employ two separate linear layers to predict the next state and reward-to-go for each timestep, respectively. The losses for each timestep are then averaged. Benefited from our bouding constraint on the state and returns-to-go together, improved results are then observed when simultaneously predicting the state and returns-to-go. 
\subsection{Diffusion based Trajectory Optimization}\label{improve}

\noindent\textbf{Notation:} To clarify the distinction between the two types of timesteps in this work, we employ a notation convention. Specifically, we use superscripts $k \in \{1, \dots, K\}$ to represent timesteps associated with the diffusion process, and subscripts $t \in \{1, \dots, T\}$ to represent timesteps pertaining to the trajectory in reinforcement learning.

The diffusion model offers the advantage of effectively modeling multi-modal data~\cite{wang2022diffusion}. To preserve this desirable characteristic, we train the diffusion model to learn the forward process of all diffusion timesteps. However, since our dataset comprises a collection of sub-optimal trajectories, the diffusion model may be affected by such sub-optimal behaviors. Hence, we incorporate condition guidance in the diffusion model. Specifically, we employ classifier-free guidance~\cite{ho2022classifier} to directly train a conditional diffusion model. This approach, in comparison to classifier guidance~\cite{dhariwal2021diffusion}, avoids the need for dynamic programming.

In the standard reinforcement learning task, our objective is to devise a policy that maximizes reward. Therefore, we employ the returns-to-go metric derived from the offline dataset as the conditioning factor denoted by $\boldsymbol{y}(\tau)$. It is noteworthy that our approach diverges from Diffuser~\cite{janner2022planning} and Decision Diffuser~\cite{ajay2022conditional}, we employ cumulative reward beyond the current timestep $t$ as the returns-to-go: $R(\tau) = \sum_{t' \geq t} \gamma^k r_{t'}$. We consider that trajectories with lower overall rewards still contain locally better trajectories. This option ensures that our diffusion model is not confined solely to the optimal trajectories. Formally, we train our conditional generative model in a supervised manner. Given a trajectory $\tau \in \mathcal{D}$ from the dataset $\mathcal{D}$, represented as $\tau = (s_t, s_{t+1}, \dots, s_T)$, it is labeled with the returns-to-go $\hat{R_t}$ which achieves starting from the current timestep.
\begin{algorithm}[t]
\renewcommand{\algorithmicrequire}{\textbf{Input:}}
\renewcommand{\algorithmicensure}{\textbf{Output:}}
\renewcommand{\algorithmiccomment}[1]{\hfill $\triangleright$ #1}
\caption{Diffusion based Trajectory Optimization}
\label{alg2:td}
\begin{algorithmic}[1]
\REQUIRE trajectory $\tau$, condition $\boldsymbol{y}(\tau)$, guidance scale $\omega$, improve step $K$, Noise model $\epsilon_\theta$ and inverse dynamics $f_\phi$.
\ENSURE improved trajectory $\tau^*$
\STATE Initialize $\boldsymbol{x}_K(\tau) \gets \text{AddNoise}(\tau)$
\FOR{$k = K \dots 1$}
    \STATE $\boldsymbol{x}_k(\tau)[0] \gets \tau[0]$ \COMMENT{Constrain plan with current state}
    \STATE  $\hat{\epsilon}\gets  (1-\omega)\epsilon_\theta(\boldsymbol{x}_k(\tau), k) +\omega\epsilon_\theta(\boldsymbol{x}_k(\tau), \boldsymbol{y}(\tau), k)$
    \STATE$(\mu_{k-1}, \sum_{k-1})\gets \text{Denoise}(\boldsymbol{x}_{k}(\tau), \hat{\epsilon})$
    \STATE $\boldsymbol{x}_{k-1}\sim \mathcal{N}(\mu_{k-1}, \sum_{k-1})$
\ENDFOR
\STATE $\tau^* \gets \boldsymbol{x}_0(\tau)$

\end{algorithmic}
\end{algorithm}
For each trajectory $\tau$ and condition $\boldsymbol{y}(\tau)$, we start by sampling noise $\epsilon$ from a standard normal distribution $\mathcal{N}(0, \boldsymbol{I})$, and a timestep $k$ uniformly from the set $\{1, \dots, K\}$. Then, we construct a noisy array of states $\boldsymbol{x}_k(\tau)$ and predict the noise as $\hat{\epsilon}_\theta = \epsilon_\theta(\boldsymbol{x}_k(\tau), \boldsymbol{y}(\tau), k)$, using the reverse diffusion process $p_\theta$ parameterized by the noise model $\theta$. The model is trained using the following loss function:
\begin{equation}
    \begin{aligned}
    \mathcal{L}(\theta) =
    \mathbb{E}_{k, \tau \in \mathcal{D}, \eta \sim \operatorname{Bern}(p)}\left[\left|\epsilon - \epsilon_{\theta}\left(\boldsymbol{x}_{k}(\tau), \hat{\boldsymbol{y}}(\tau, \eta), k\right)\right|^{2}\right],
\end{aligned}
\end{equation}
Here, $\hat{\boldsymbol{y}}(\tau, \eta) = (1-\eta) \boldsymbol{y}(\tau)+\eta \emptyset $, $\operatorname{Bern}(p)$ represents a Bernoulli distribution with probability $p$, defined as:
\begin{equation}
    \text{Bern}(p)=\left\{\begin{matrix}
  1 ,&p\\
  0 ,&1-p
\end{matrix}\right .
\end{equation}

During sampling, we optimize our trajectory through the reverse process of a conditional diffusion model, given by
\begin{equation}
    \begin{aligned}
    \pi_\theta(\boldsymbol{x}_0(\tau)|\boldsymbol{y}(\tau)) &= p_\theta(\boldsymbol{x}_{0:k}(\tau)|\boldsymbol{y}(\tau)),
\end{aligned}
\end{equation}
The final trajectory of the reverse chain, $\boldsymbol{x}_0(\tau)$, represents the improved trajectory. Generally, $p_\theta(\boldsymbol{x}_{k-1}|\boldsymbol{x}_k, \boldsymbol{y}(\tau))$ can be modeled as a Gaussian distribution:
\begin{equation}
\begin{aligned}
    p_\theta(\boldsymbol{x}_{k-1}|\boldsymbol{x}_k, \boldsymbol{y}(\tau)) = \mathcal{N}(\boldsymbol{x}_{k-1}; \mu_\theta(\boldsymbol{x}_k, \boldsymbol{y}(\tau), k), \Sigma_\theta(\boldsymbol{x}_k, \boldsymbol{y}(\tau), k)),
\end{aligned}
\end{equation}
We follow~\cite{ho2020denoising} to parameterize $p_\theta(\boldsymbol{x}_{k-1}(\tau)|\boldsymbol{x}_k(\tau))$ as a noise prediction model, with the covariance matrix fixed as $\Sigma_\theta(\boldsymbol{x}_k(\tau), \boldsymbol{y}(\tau), k) = \beta_k\boldsymbol{I}$, and the mean constructed as:
\begin{equation}
    \begin{aligned}
    \mu_\theta(\boldsymbol{x}_k(\tau), \boldsymbol{y}(\tau), k) = \frac{1}{\sqrt{\alpha_k}}\left(\boldsymbol{x}_k(\tau) - \frac{\beta_k}{\sqrt{1-\bar{\alpha}_k}}\hat{\epsilon}_\theta(\boldsymbol{x}_k(\tau), \boldsymbol{y}(\tau), k)\right),
\end{aligned}
\end{equation}
where $\hat{\epsilon}_\theta(\boldsymbol{x}_k(\tau), \boldsymbol{y}(\tau), k))$ represents perturbed noise, following~\cite{ho2022classifier}. It is calculated as:
\begin{equation}
\begin{aligned}
    \hat{\epsilon}_\theta(\boldsymbol{x}_k(\tau), \boldsymbol{y}(\tau), k)) = \epsilon_{\theta}\left(\boldsymbol{x}_{k}(\tau), \emptyset, k\right) + \omega\left(\epsilon_{\theta}\left(\boldsymbol{x}_{k}(\tau), \boldsymbol{y}(\tau), k\right) - \epsilon_{\theta}\left(\boldsymbol{x}_{k}(\tau), \emptyset, k\right)\right),
\end{aligned}
\end{equation}
The scalar $\omega$ is applied to the term $\epsilon_{\theta}\left(\boldsymbol{x}_{k}(\tau), \boldsymbol{y}(\tau), k\right)- \epsilon_{\theta}\left(\boldsymbol{x}_{k}(\tau), \emptyset, k\right)$, aiming to enhance and extract the most valuable parts of trajectories in the dataset that exhibit $y(\tau)$. 

With these components, trajectory optimization becomes similar to sampling from the diffusion model. First, we obtain a trajectory $\tau$, represented as $\boldsymbol{x}_k(\tau)$ with $k \ll K$. Next, we sample states starting from timestep $k$ using the diffusion process conditioned on $\boldsymbol{y}(\tau)$. Finally, we determine the action that should be taken to reach the predicted state with our inverse dynamics model. The process is iterated within a conventional receding-horizon control loop, as outlined in \Cref{alg2:td}.

\begin{figure}[t]
    \centering
    \includegraphics[width=\linewidth]{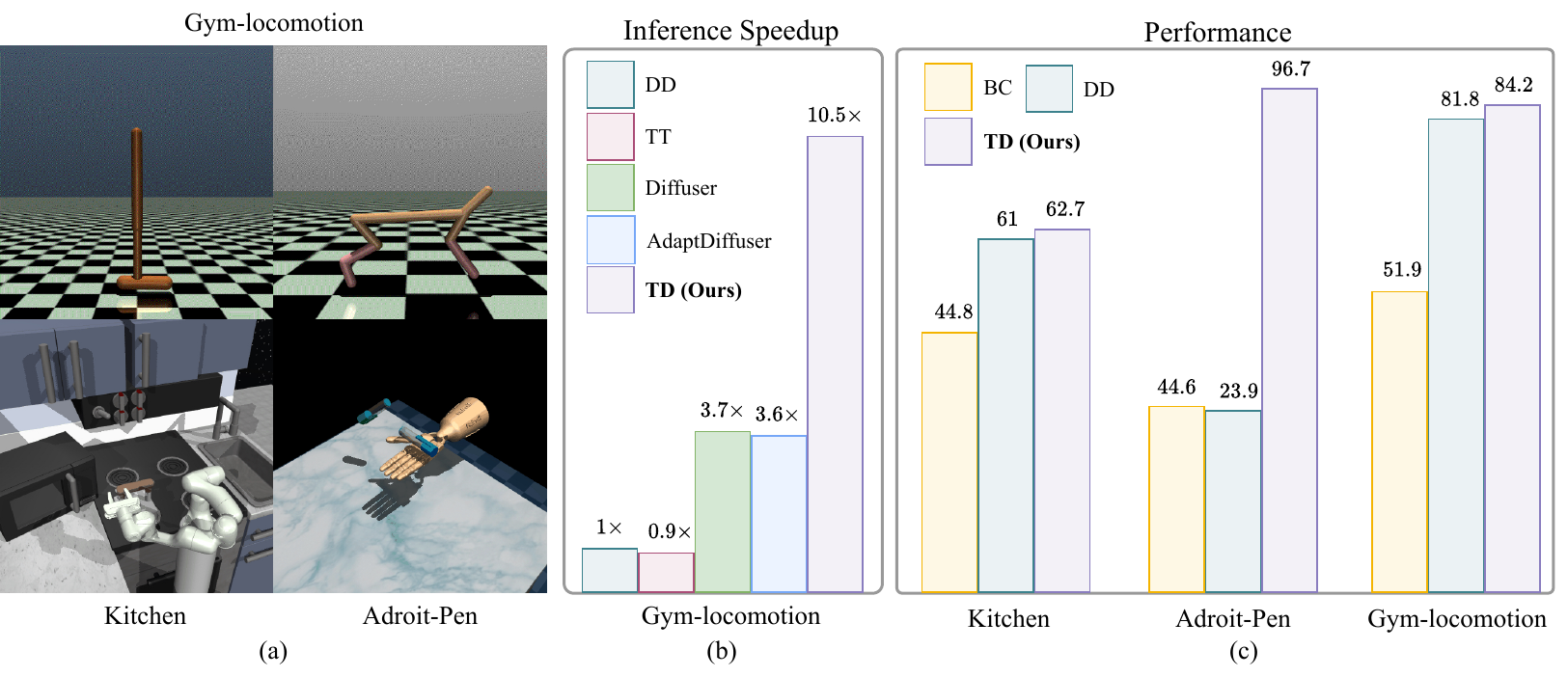}
    \caption{\textbf{Tasks and Results Overview.} (a) We study the following tasks: Gym-locomotion, Franka Kitchen and Adroit-Pen. (b) Compared to the previous naive diffusion model, Trajectory Diffuser achieves remarkable improvements in terms of inference speed. (c) Our method outperforms prior sequence modeling methods such as Decision Diffuser and Behavioral Cloning (BC) method. For performance metrics, we use normalized average returns~\cite{fu2020d4rl} for D4RL tasks.}
    \label{fig:speed}
\end{figure}

\section{Experiments}
\label{sec:experiments}
In the experiment, conclusions are drawn that our method can improve inference speed without sacrificing performance as summarized in \Cref{fig:speed}. We also conduct ablation experiments to demonstrate the necessity of trajectory improvement and the tradeoff between performance and efficiency. Specifically, we aim to answer the following questions:

\begin{itemize}
    \item How much speed improvement can be achieved by Trajectory Diffuser compared to prior naive diffusion model based methods?
    \item Does generating the feasible trajectory using an autoregressive model affect the model's performance?
    \item Whether our method is applicable to other regression models?
    \item Is the process of trajectory improvement necessary? Where is the balance point between efficiency and performance?
    
\end{itemize}

\noindent\textbf{Dataset.} We use the D4RL dataset~\cite{fu2020d4rl}, which provides benchmark datasets for reinforcement learning. We consider three different domains of tasks in D4RL, including Gym-locomotion, Adroit, and Kitchen. 


\subsection{Evaluation on the Efficiency}
\begin{table}[b]
    \caption{\textbf{Testing Time in D4RL MuJoCo Environment.} All these data are tested with a single \textit{NVIDIA A100 GPU}. The unit in the table is second (s).}
    \small
    \centering
    \begin{tabular}{c|ccc|c}
    \toprule[1.2pt]
        \textbf{Dataset} & \multicolumn{3}{c|}{Med-Expert}& \multirow{2}{*}{\textbf{Average SPS}}  \\ \cmidrule{1-4}
        \textbf{Environment} & HalfCheetah & Hopper & Walker2d & \\ \midrule
        TT & 8.43 s & 2.68 s & 5.17 s & 5.43 s \\ 
        Diffuser & 1.17 s & 1.26 s & 1.28 s & 1.24 s \\ 
        DD & 4.26 s & 4.47 s & 5.11 s & 4.61 s\\ 
        AdaptDiffuser & 1.21 s & 1.29 s & 1.30 s & 1.27 s\\
        TD(ours) & \textbf{0.27} s & \textbf{0.42} s & \textbf{0.62} s & \textbf{0.44} s\\
    \bottomrule[1.2pt]
    \end{tabular}
    
    \label{table:time}
\end{table}

The inference time of the model in the control domain is also an important issue. Previous sequence models based on diffusion achieve significant improvements in performance. However, in each inference of the model, a reverse process of the diffusion model is performed, which significantly increases the inference time. In this section, we focus on evaluating the efficiency of Trajectory Diffuser.

\noindent\textbf{Experimental Setup.} In order to ensure the validity of the comparisons, we conduct a comparative analysis among four trajectory planning models, including Diffuser~\cite{janner2022planning}, Trajectory Transformer~(TT)~\cite{janner2021offline}, Decision Diffuser~(DD)~\cite{ajay2022conditional} and AdaptDiffuser\cite{liang2023adaptdiffuser}. The planning horizon and diffusion step for the trajectory planning are consistent with the hyperparameters provided in the original paper. We choose Gym-locomotion task as the testbed, we sample $1000$ transitions by interacting with the environment and calculate the corresponding seconds-per-step~(SPS) for inference speed.

\noindent\textbf{Results.} We show the inference time of generating an action taken by prior methods and our method in \Cref{table:time}.  Taking DD as the baseline, we are able to attribute the performance boost to specific techniques proposed, as visualized in \Cref{fig:speed}. Specifically, compared to the transformer-based TT model, the TT model uses beam search to obtain the optimal trajectory, while our model utilizes a diffusion model to optimize the trajectory, resulting in more than $10\times$ increase in inference speed. Compared to naive diffusion models including DD, Diffuser, and AdaptDiffuser, our method achieves an inference speedup of $3$-$10\times$.

\subsection{Evaluation on the Effectiveness}
\begin{table*}[t]
    \caption{\textbf{The performance of Trajectory Diffuser and a variety of sequence modeling algorithms on the D4RL locomotion benchmark}~\cite{fu2020d4rl}. Results for Trajectory Diffuser correspond to the mean and standard error over 20 planning seeds. We emphasize in bold scores within 5 percent of the maximum per task ($\ge$ 0.95$\cdot$max).}
    \small
    \centering
    \resizebox{\linewidth}{!}{
    \begin{tabular}{llccccccc}
    \toprule[1.2pt]
        \textbf{Dataset} & \textbf{Environment} & \textbf{BC} & \textbf{DT} & \textbf{TT}  & \textbf{Diffuser} & \textbf{DD} & \textbf{AdaptDiffuser} & \textbf{TD~(Ours)} \\ \midrule
        Med-Expert & HalfCheetah & 55.2   & 86.8  $\pm$1.3   & \textbf{95.0}   $\pm$0.2    & 88.9 $\pm$0.3  & \textbf{90.6}  $\pm$1.3  & 89.6  $\pm$0.8  & \textbf{92.2}  $\pm$0.8 \\ 
        Med-Expert & Hopper      & 52.5   & 107.6 $\pm$1.8  & 110.0   $\pm$2.7   & 103.3 $\pm$1.3 & \textbf{111.8}  $\pm$1.8 & \textbf{111.6} $\pm$2.0 & \textbf{113.6} $\pm$1.5 \\ 
        Med-Expert & Walker2d    & \textbf{107.5}  & \textbf{108.1} $\pm$0.2  & 101.9 $\pm$6.8 & \textbf{106.9}   $\pm$0.2 & \textbf{108.8}  $\pm$1.7 & \textbf{108.2} $\pm$0.8 & \textbf{109.0} $\pm$0.6 \\ \midrule
        Medium & HalfCheetah     & 42.6   & 42.6  $\pm$0.1   & \textbf{46.9} $\pm$0.4  & 42.8   $\pm$0.3  & \textbf{49.1}  $\pm$1.0  & 44.2  $\pm$0.6  & 44.2  $\pm$1.1 \\ 
        Medium & Hopper          & 52.9   & 67.6  $\pm$1.0   & 61.1 $\pm$3.6  & 74.3   $\pm$1.4  & 79.3  $\pm$3.6  & \textbf{96.6}  $\pm$2.7  & \textbf{97.1}  $\pm$1.5 \\ 
        Medium & Walker2d        & 75.3   & 74.0  $\pm$1.4   & 79.0   $\pm$2.8   & 79.6  $\pm$0.5  & \textbf{82.5}  $\pm$1.4  & \textbf{84.4}  $\pm$2.6  & \textbf{83.5}  $\pm$2.2 \\ \midrule
        Med-Replay & HalfCheetah & 36.6   & 36.6  $\pm$0.8   & \textbf{41.9} $\pm$2.5  & 37.3   $\pm$0.5  & 39.3  $\pm$4.1  & 38.3  $\pm$0.9  & \textbf{41.3}  $\pm$0.8 \\ 
        Med-Replay & Hopper      & 18.1   & 82.7  $\pm$7.0   & 91.5 $\pm$3.6  & 93.6   $\pm$0.4  & \textbf{100.0}   $\pm$0.7   & 92.2  $\pm$1.5  & 94.3  $\pm$3.2 \\ 
        Med-Replay & Walker2d    & 26.0     & 66.6  $\pm$3.0   & \textbf{82.6}  $\pm$6.9  & 70.6   $\pm$1.6  & 75.0    $\pm$4.3   & \textbf{84.7}  $\pm$3.1  & \textbf{82.5}  $\pm$2.6 \\ \midrule
        \multicolumn{2}{c}{\textbf{Average Returns}} & 51.9 & 74.7 & 78.9  & 77.5  & 81.8 & 83.4  & \textbf{84.2}\\ \midrule
        \multicolumn{2}{c}{\textbf{Runtime Per Action }} & - & - & 5.43~s  & 1.24~s  & 4.61~s & 1.27~s  & \textbf{0.44}~s\\
    \bottomrule[1.2pt]
    \end{tabular}
    }
    
    \label{table:result 1}
\end{table*}

\begin{table}[b]
    \caption{\textbf{Comparison on Adroit tasks and Kitchen tasks.}}
    \small
    \centering
    \begin{tabular}{lllccc}

    \toprule[1.2pt]
        \textbf{Dataset} & \textbf{Env} & \textbf{BC} & \textbf{CQL}  & \textbf{DD} & \textbf{TD~(Ours)} \\ \midrule 
        Human  & Pen & 25.8  & 35.2 & 7.5 $\pm$13.6 & \textbf{87.9} $\pm$63.1  \\ 
        Expert  & Pen & 69.7 & -  & 9.0 $\pm$29.8 & \textbf{116.1} $\pm$54.6 \\ 
        Cloned  & Pen & 38.3 & 27.2  & 55.1 $\pm$66.2 & \textbf{86.1} $\pm$57.9\\
        \midrule[1.2pt]
        \multicolumn{2}{c}{\textbf{Average}} & 44.6 & - & 23.9 & \textbf{96.7} \\
        \midrule[1.2pt]
        Partial & Kitchen & 38.0 & 49.8 & 57.0 $\pm$2.8& \textbf{71.2} $\pm$11.0 \\
        Mixed & Kitchen & 51.5 & 51.0 & \textbf{65.0} $\pm$2.5 & 54.2 $\pm$9.3  \\ \midrule
        \multicolumn{2}{c}{\textbf{Average}}  & 44.8 & 50.4 & 61.0 & \textbf{62.7} \\
        \bottomrule[1.2pt]
    \end{tabular}
    
    \label{table:Adroit and Kitchen}
\end{table}

We are interested in investigating whether the Trajectory Diffuser achieves inference speed improvements while potentially causing a decline in model performance. Thus we conduct an evaluation of Trajectory Diffuser in Gym-locomotion, Adroit and Kitchen tasks and compare its performance against previous sequence-modeling methods. Gym-locomotion results are shown in \Cref{table:result 1}. The results of Adroit and Kitchen tasks are shown in \Cref{table:Adroit and Kitchen}. Additionally, in the \textcolor{red!60!black}{Appendix}, we compare our approach with other methods based on Temporal Difference Learning and model-based methods.

\noindent\textbf{Baselines.} we compare our approach with Behavior Cloning (BC) and sequence modeling methods include the Decision Transformer~\cite{chen2021decision}, Trajectory Transformer~\cite{janner2021offline}, diffusion-based models include Diffuser~\cite{janner2022planning}, Decision Diffusion~\cite{ajay2022conditional} and AdaptDiffuser~\cite{liang2023adaptdiffuser}. To provide a comprehensive evaluation of our method's performance, in Appendix, we compare our method with several existing offline RL methods, including CQL~\cite{kumar2020conservative}, BRAC~\cite{wu2019behavior}, IQL~\cite{kostrikov2021offline}, Diffusion-QL~\cite{wang2022diffusion}, and MoReL~\cite{kidambi2020morel}.

\noindent\textbf{Experimental Setup.} We independently train a sequence model based on the transformer architecture, incorporating a state diffusion process and an inverse dynamics model. This training is carried out using publicly available D4RL datasets. We train the model for 200k iterations (100k iterations for Adroit tasks and Kitchen tasks) with a batch size of 32. The specific hyperparameters varied across different datasets and can be found in the supplementary material.

\noindent\textbf{Results.} The experimental results in \Cref{table:result 1} and \Cref{table:Adroit and Kitchen} demonstrate that our method achieves state-of-the-art performance among sequence-based models across a diverse range of offline reinforcement learning tasks. Particularly, in the more challenging D4RL Adroit and Kitchen tasks, which require strong policy regularization to overcome extrapolation errors~\cite{fujimoto2019off} and complex credit assignment over long periods, our method outperforms Decision Diffusion and exhibits significant advantages. 

\subsection{Compare with More Autoregressive Models}

Our method introduces the utilization of autoregressive models for generating feasible trajectories. Traditional autoregressive models, such as GRU\cite{chung2014empirical}, LSTM\cite{shi2015convolutional}, and Transformer\cite{vaswani2017attention}, are commonly employed in this domain. Through the incorporation of diverse autoregressive models, we establish the robust generalization capabilities of our proposed framework. Furthermore, we empirically validate that leveraging the Transformer model for trajectory initialization yields superior outcomes within our approach.
\begin{table}[!h]
    \caption{Compare with different autoregressive models, evaluation on Medium-Expert Dataset. (-) represent without trajectory optimization}
    \label{table:more regressive}
    \centering
    \small
    \resizebox{0.9\linewidth}{!}{
    \begin{tabular}{l|ccc|ccc}
    \toprule[1.2pt]
        \textbf{Env} & \textbf{Transformer} & \textbf{GRU} & \textbf{LSTM} & \textbf{Transformer~(-)} & \textbf{GRU~(-)} & \textbf{LSTM~(-)} \\ 
        \midrule
        HalfCheetah & \textbf{92.2$\pm$0.8} & 80.8$\pm$1.2 & 83.1$\pm$1.7 &\textbf{86.2 $\pm$1.2}& 75.8$\pm$7.1 & 77.6$\pm$9.9 \\ 
        Hopper & \textbf{113.6$\pm$1.5} & 111.9$\pm$0.9 & 109.2$\pm$1.0 &\textbf{103.4 $\pm$4.7} & 101.5$\pm$11.3 & 90.1$\pm$23.4 \\ 
        Walker2d &\textbf{109.0$\pm$0.6} & 104.8$\pm$8.4 & 107.8$\pm$1.4 &103.3 $\pm$6.6 & 99.4$\pm$5.2 & \textbf{105.9$\pm$2.7} \\
    \bottomrule[1.2pt]
    \end{tabular}
    }
\end{table}\\
\noindent\textbf{Results.} The results shown in \Cref{table:more regressive} demonstrate that the trajectories generated with the assistance of Transformer, which excels at modeling historical information, exhibit superior quality. Our approach is capable of extending to different autoregressive models, and employing the diffusion model for trajectory optimization leads to a significant enhancement in trajectory quality.

\subsection{Ablation Study of Trajectory Diffuser}

\begin{table}[t]
    \caption{Ablation study on trajectory optimizer. (-) represent without optimization.}

    \small
    \centering
    \resizebox{\linewidth}{!}{
    \begin{tabular}{l|ccc|ccc|ccc|c}
    \toprule[1.2pt]
        \textbf{Dataset} & \multicolumn{3}{c|}{Med-Expert}& \multicolumn{3}{c|}{Medium}& \multicolumn{3}{c|}{Med-Replay}& \multirow{2}{*}{\textbf{Average}}  \\ \cmidrule{1-10}
        \textbf{Env} & HalfCheetah & Hopper & Walker2d & HalfCheetah & Hopper & Walker2d & HalfCheetah & Hopper & Walker2d & \\ \midrule
        \textbf{TD~(-)} & 86.2 $\pm$1.2 & 103.4 $\pm$4.7 & 103.3 $\pm$6.6 & 40.0 $\pm$8.3 & \textbf{98.8}  $\pm$1.5 & 79.0 $\pm$10.0 & 38.5 $\pm$2.5 & 79.1 $\pm$9.1 & 75.5  $\pm$3.5 & 79.1 \\ 
        \textbf{TD} & \textbf{92.2}  $\pm$0.8 & \textbf{113.6}  $\pm$1.5 & \textbf{109.3}  $\pm$0.8 & \textbf{44.2} $\pm$1.1 & 97.1 $\pm$1.5 & \textbf{83.5}  $\pm$2.2 & \textbf{41.3}  $\pm$0.8 & \textbf{94.3} $\pm$3.2 & \textbf{82.5}  $\pm$2.6 & \textbf{84.2}\\

    \bottomrule[1.2pt]
    \end{tabular}
    }
    
    \label{table:result 2}
\end{table}
We provide empirical evidence in support of utilizing a diffusion model for trajectory enhancement through rigorous experimentation. By conducting ablation experiments, we evaluate the influence of various optimization steps on performance, clearly demonstrating that the optimization process primarily takes place during the last tenth of the diffusion model. Furthermore, comprehensive information regarding the tuning range of hyperparameters is documented in the Appendix to offer a thorough understanding on the advantage of our model.\\
\noindent\textbf{Results.} As shown in \Cref{table:result 2}, the diffusion model improves the performance and reduces the variance of the model. \Cref{table:result 3} indicates that performance is minimally affected when using only one-tenth of the usual number of diffusion steps. However, with longer diffusion steps, performance declines. This suggests that with increasing optimization steps, the injection of noise corresponding to the $k$ steps corrupts the well-initialized trajectories in the first stage, bringing the model's performance closer to that of the pure diffusion model.

\begin{table}[ht]

    \caption{Ablation study on different diffusion steps.}
    \small
    \centering
    \resizebox{0.6\linewidth}{!}{
    \begin{tabular}{c|ccc|c}
    \toprule[1.2pt]
        \textbf{Dataset} & \multicolumn{3}{c|}{Med-Expert}& \multirow{2}{*}{\textbf{Average}}  \\ \cmidrule{1-4}
        \textbf{Env} & HalfCheetah & Hopper & Walker2d & \\ \midrule
        2 steps & \textbf{92.2} $\pm$0.8 & 110.2 $\pm$2.7 & 108.7 $\pm$1.6 & 103.7 \\ 
        5 steps & 92.1 $\pm$1.0 & \textbf{113.6} $\pm$1.5 & \textbf{109.0} $\pm$0.8 & \textbf{105.0} \\ 
        10 steps & 88.4  $\pm$1.4 & 111.7 $\pm$1.9 & 108.9 $\pm$2.1 & 103.0\\ 
        20 steps & 83.9 $\pm$6.4 & 103.1 $\pm$2.6 & 108.3 $\pm$0.3 & 98.4\\
        50 steps & 83.3 $\pm$8.5 & 100.2 $\pm$2.5 & 108.4 $\pm$0.3 & 97.3\\
        100 steps & 82.3 $\pm$19.5 & 106.2 $\pm$1.9 & 108.3 $\pm$0.4 & 98.9 \\
    \bottomrule[1.2pt]
    \end{tabular}
    }
    
    \label{table:result 3}
\end{table}

\section{Conclusion}
\label{sec:conclusion}
In this paper, we propose a decomposition approach for the inference process of diffusion models, consisting of generating feasible trajectories and optimizing them. By utilizing a more efficient autoregressive model for trajectory generation and leveraging the capabilities of the diffusion model for trajectory optimization, we achieve significant improvements in inference speed while maintaining high-quality results. The experimental results demonstrate the impressive speed improvements of our method compared to traditional diffusion models. Additionally, our approach outperforms prior sequence modeling methods in terms of effectiveness. Benefiting from the improvement in both efficiency and accuracy, we look forward to our approach as a practical framework for decision-making tasks, and also an example design on how to apply the diffusion model for solving offline reinforcement learning problems.

\subsubsection{Acknowledgements.}
This work was supported in part by the National Natural Science Foundation of China under grant U23B2011, 62102069, U20B2063 and 62220106008, the Key R\&D Program of Zhejiang under grant 2024SSYS0091, the Sichuan Science and Technology Program under Grant 2024NSFTD0034

%
%
\bibliographystyle{splncs04}
\bibliography{main}

\begin{thebibliography}{10}
\providecommand{\url}[1]{\texttt{#1}}
\providecommand{\urlprefix}{URL }
\providecommand{\doi}[1]{https://doi.org/#1}

\bibitem{agrawal2016learning}
Agrawal, P., Nair, A.V., Abbeel, P., Malik, J., Levine, S.: Learning to poke by poking: Experiential learning of intuitive physics. Advances in neural information processing systems  \textbf{29} (2016)

\bibitem{ajay2022conditional}
Ajay, A., Du, Y., Gupta, A., Tenenbaum, J., Jaakkola, T., Agrawal, P.: Is conditional generative modeling all you need for decision-making? arXiv preprint arXiv:2211.15657  (2022)

\bibitem{carion2020end}
Carion, N., Massa, F., Synnaeve, G., Usunier, N., Kirillov, A., Zagoruyko, S.: End-to-end object detection with transformers. In: European conference on computer vision. pp. 213--229. Springer (2020)

\bibitem{chen2021decision}
Chen, L., Lu, K., Rajeswaran, A., Lee, K., Grover, A., Laskin, M., Abbeel, P., Srinivas, A., Mordatch, I.: Decision transformer: Reinforcement learning via sequence modeling. Advances in neural information processing systems  \textbf{34},  15084--15097 (2021)

\bibitem{chung2014empirical}
Chung, J., Gulcehre, C., Cho, K., Bengio, Y.: Empirical evaluation of gated recurrent neural networks on sequence modeling. arXiv preprint arXiv:1412.3555  (2014)

\bibitem{devlin2018bert}
Devlin, J., Chang, M.W., Lee, K., Toutanova, K.: Bert: Pre-training of deep bidirectional transformers for language understanding. arXiv preprint arXiv:1810.04805  (2018)

\bibitem{dhariwal2021diffusion}
Dhariwal, P., Nichol, A.: Diffusion models beat gans on image synthesis. Advances in neural information processing systems  \textbf{34},  8780--8794 (2021)

\bibitem{du2019implicit}
Du, Y., Mordatch, I.: Implicit generation and generalization in energy-based models. arXiv preprint arXiv:1903.08689  (2019)

\bibitem{fu2020d4rl}
Fu, J., Kumar, A., Nachum, O., Tucker, G., Levine, S.: D4rl: Datasets for deep data-driven reinforcement learning. arXiv preprint arXiv:2004.07219  (2020)

\bibitem{fujimoto2019off}
Fujimoto, S., Meger, D., Precup, D.: Off-policy deep reinforcement learning without exploration. In: International conference on machine learning. pp. 2052--2062. PMLR (2019)

\bibitem{furuta2021generalized}
Furuta, H., Matsuo, Y., Gu, S.S.: Generalized decision transformer for offline hindsight information matching. arXiv preprint arXiv:2111.10364  (2021)

\bibitem{grathwohl2020learning}
Grathwohl, W., Wang, K.C., Jacobsen, J.H., Duvenaud, D., Zemel, R.: Learning the stein discrepancy for training and evaluating energy-based models without sampling. In: International Conference on Machine Learning. pp. 3732--3747. PMLR (2020)

\bibitem{hafner2019learning}
Hafner, D., Lillicrap, T., Fischer, I., Villegas, R., Ha, D., Lee, H., Davidson, J.: Learning latent dynamics for planning from pixels. In: International conference on machine learning. pp. 2555--2565. PMLR (2019)

\bibitem{he2022masked}
He, K., Chen, X., Xie, S., Li, Y., Doll{\'a}r, P., Girshick, R.: Masked autoencoders are scalable vision learners. In: Proceedings of the IEEE/CVF conference on computer vision and pattern recognition. pp. 16000--16009 (2022)

\bibitem{ho2020denoising}
Ho, J., Jain, A., Abbeel, P.: Denoising diffusion probabilistic models. Advances in neural information processing systems  \textbf{33},  6840--6851 (2020)

\bibitem{ho2022classifier}
Ho, J., Salimans, T.: Classifier-free diffusion guidance. arXiv preprint arXiv:2207.12598  (2022)

\bibitem{janner2022planning}
Janner, M., Du, Y., Tenenbaum, J.B., Levine, S.: Planning with diffusion for flexible behavior synthesis. arXiv preprint arXiv:2205.09991  (2022)

\bibitem{janner2021offline}
Janner, M., Li, Q., Levine, S.: Offline reinforcement learning as one big sequence modeling problem. Advances in neural information processing systems  \textbf{34},  1273--1286 (2021)

\bibitem{kidambi2020morel}
Kidambi, R., Rajeswaran, A., Netrapalli, P., Joachims, T.: Morel: Model-based offline reinforcement learning. Advances in neural information processing systems  \textbf{33},  21810--21823 (2020)

\bibitem{kinga2015method}
Kinga, D., Adam, J.B., et~al.: A method for stochastic optimization. In: International conference on learning representations (ICLR). vol.~5, p.~6. San Diego, California; (2015)

\bibitem{kostrikov2021offline}
Kostrikov, I., Fergus, R., Tompson, J., Nachum, O.: Offline reinforcement learning with fisher divergence critic regularization. In: International Conference on Machine Learning. pp. 5774--5783. PMLR (2021)

\bibitem{kumar2020conservative}
Kumar, A., Zhou, A., Tucker, G., Levine, S.: Conservative q-learning for offline reinforcement learning. Advances in Neural Information Processing Systems  \textbf{33},  1179--1191 (2020)

\bibitem{levine2020offline}
Levine, S., Kumar, A., Tucker, G., Fu, J.: Offline reinforcement learning: Tutorial, review. and Perspectives on Open Problems  \textbf{5} (2020)

\bibitem{liang2023adaptdiffuser}
Liang, Z., Mu, Y., Ding, M., Ni, F., Tomizuka, M., Luo, P.: Adaptdiffuser: Diffusion models as adaptive self-evolving planners. arXiv preprint arXiv:2302.01877  (2023)

\bibitem{mish1908self}
Mish, M.D.: A self regularized non-monotonic neural activation function. arxiv 2019. arXiv preprint arXiv:1908.08681  (1908)

\bibitem{nagabandi2018neural}
Nagabandi, A., Kahn, G., Fearing, R.S., Levine, S.: Neural network dynamics for model-based deep reinforcement learning with model-free fine-tuning. In: 2018 IEEE international conference on robotics and automation (ICRA). pp. 7559--7566. IEEE (2018)

\bibitem{ni2023metadiffuser}
Ni, F., Hao, J., Mu, Y., Yuan, Y., Zheng, Y., Wang, B., Liang, Z.: Metadiffuser: Diffusion model as conditional planner for offline meta-rl. arXiv preprint arXiv:2305.19923  (2023)

\bibitem{nichol2021glide}
Nichol, A., Dhariwal, P., Ramesh, A., Shyam, P., Mishkin, P., McGrew, B., Sutskever, I., Chen, M.: Glide: Towards photorealistic image generation and editing with text-guided diffusion models. arXiv preprint arXiv:2112.10741  (2021)

\bibitem{nichol2021improved}
Nichol, A.Q., Dhariwal, P.: Improved denoising diffusion probabilistic models. In: International Conference on Machine Learning. pp. 8162--8171. PMLR (2021)

\bibitem{nijkamp2019learning}
Nijkamp, E., Hill, M., Zhu, S.C., Wu, Y.N.: Learning non-convergent non-persistent short-run mcmc toward energy-based model. Advances in Neural Information Processing Systems  \textbf{32} (2019)

\bibitem{pathak2018zero}
Pathak, D., Mahmoudieh, P., Luo, G., Agrawal, P., Chen, D., Shentu, Y., Shelhamer, E., Malik, J., Efros, A.A., Darrell, T.: Zero-shot visual imitation. In: Proceedings of the IEEE conference on computer vision and pattern recognition workshops. pp. 2050--2053 (2018)

\bibitem{puterman2014markov}
Puterman, M.L.: Markov decision processes: discrete stochastic dynamic programming. John Wiley \& Sons (2014)

\bibitem{radford2018improving}
Radford, A., Narasimhan, K., Salimans, T., Sutskever, I., et~al.: Improving language understanding by generative pre-training  (2018)

\bibitem{radford2019language}
Radford, A., Wu, J., Child, R., Luan, D., Amodei, D., Sutskever, I., et~al.: Language models are unsupervised multitask learners. OpenAI blog  \textbf{1}(8), ~9 (2019)

\bibitem{ramesh2022hierarchical}
Ramesh, A., Dhariwal, P., Nichol, A., Chu, C., Chen, M.: Hierarchical text-conditional image generation with clip latents. arXiv preprint arXiv:2204.06125  \textbf{1}(2), ~3 (2022)

\bibitem{ritter2020rapid}
Ritter, S., Faulkner, R., Sartran, L., Santoro, A., Botvinick, M., Raposo, D.: Rapid task-solving in novel environments. arXiv preprint arXiv:2006.03662  (2020)

\bibitem{saharia2022photorealistic}
Saharia, C., Chan, W., Saxena, S., Li, L., Whang, J., Denton, E.L., Ghasemipour, K., Gontijo~Lopes, R., Karagol~Ayan, B., Salimans, T., et~al.: Photorealistic text-to-image diffusion models with deep language understanding. Advances in Neural Information Processing Systems  \textbf{35},  36479--36494 (2022)

\bibitem{shi2015convolutional}
Shi, X., Chen, Z., Wang, H., Yeung, D.Y., Wong, W.K., Woo, W.c.: Convolutional lstm network: A machine learning approach for precipitation nowcasting. Advances in neural information processing systems  \textbf{28} (2015)

\bibitem{sohl2015deep}
Sohl-Dickstein, J., Weiss, E., Maheswaranathan, N., Ganguli, S.: Deep unsupervised learning using nonequilibrium thermodynamics. In: International conference on machine learning. pp. 2256--2265. PMLR (2015)

\bibitem{song2020denoising}
Song, J., Meng, C., Ermon, S.: Denoising diffusion implicit models. arXiv preprint arXiv:2010.02502  (2020)

\bibitem{vaswani2017attention}
Vaswani, A., Shazeer, N., Parmar, N., Uszkoreit, J., Jones, L., Gomez, A.N., Kaiser, {\L}., Polosukhin, I.: Attention is all you need. Advances in neural information processing systems  \textbf{30} (2017)

\bibitem{wang2022diffusion}
Wang, Z., Hunt, J.J., Zhou, M.: Diffusion policies as an expressive policy class for offline reinforcement learning. arXiv preprint arXiv:2208.06193  (2022)

\bibitem{wu2019behavior}
Wu, Y., Tucker, G., Nachum, O.: Behavior regularized offline reinforcement learning. arXiv preprint arXiv:1911.11361  (2019)

\bibitem{wu2018group}
Wu, Y., He, K.: Group normalization. In: Proceedings of the European conference on computer vision (ECCV). pp. 3--19 (2018)

\bibitem{zambaldi2018deep}
Zambaldi, V., Raposo, D., Santoro, A., Bapst, V., Li, Y., Babuschkin, I., Tuyls, K., Reichert, D., Lillicrap, T., Lockhart, E., et~al.: Deep reinforcement learning with relational inductive biases. In: International conference on learning representations (2018)

\bibitem{zhang2023saformer}
Zhang, Q., Zhang, L., Xu, H., Shen, L., Wang, B., Chang, Y., Wang, X., Yuan, B., Tao, D.: Saformer: A conditional sequence modeling approach to offline safe reinforcement learning. arXiv preprint arXiv:2301.12203  (2023)

\end{thebibliography}

\onecolumn
\newpage
\appendix
\section{Appendix}
\renewcommand\thefigure{A\arabic{figure}}
\renewcommand\thetable{A\arabic{table}} 
\setcounter{algorithm}{0}
\setcounter{figure}{0}
\setcounter{table}{0}
\subsection{Analyse From the Frequency Perspective}
\begin{figure}[h]
    \centering
    \includegraphics[width=\linewidth]{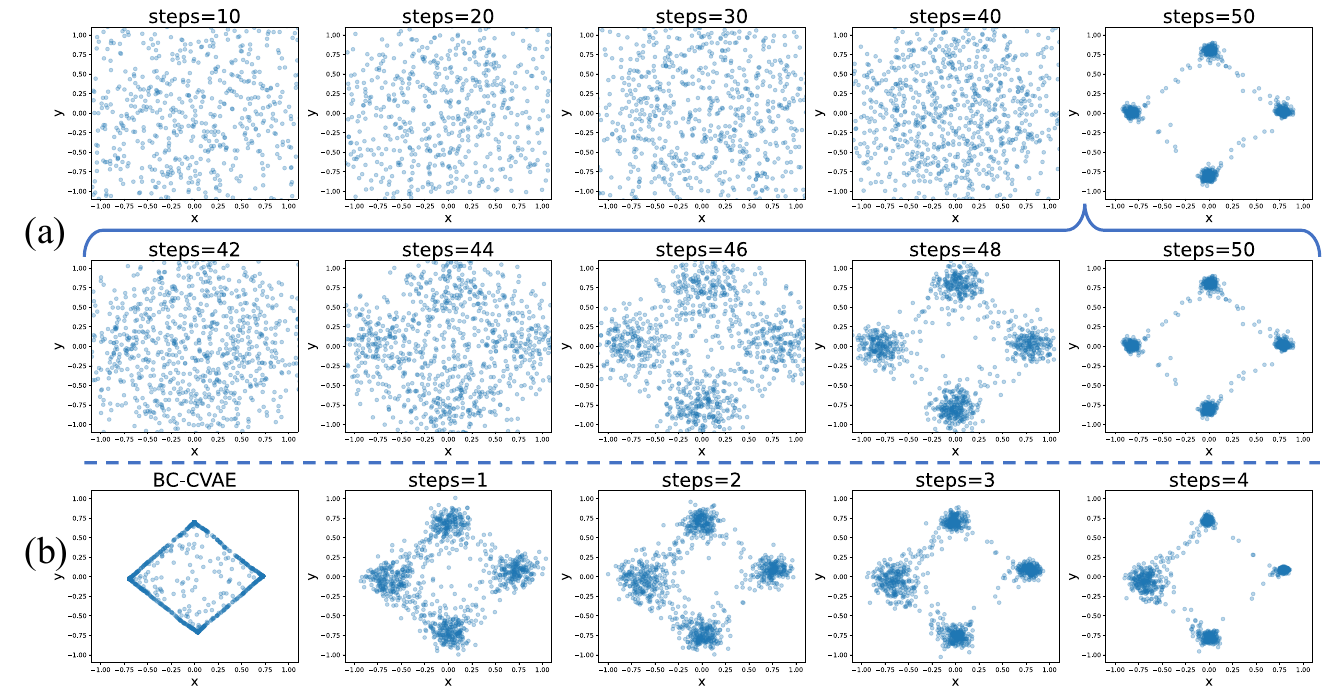}
    \caption{Bandit case. (a)~Unfolding of the diffusion process. (b)~Diffusion models as optimizers to enhance \textit{BC-CVAE}.}
    \label{fig1}
\end{figure}

\begin{figure}[t]
    \centering
    \includegraphics[width=\linewidth]{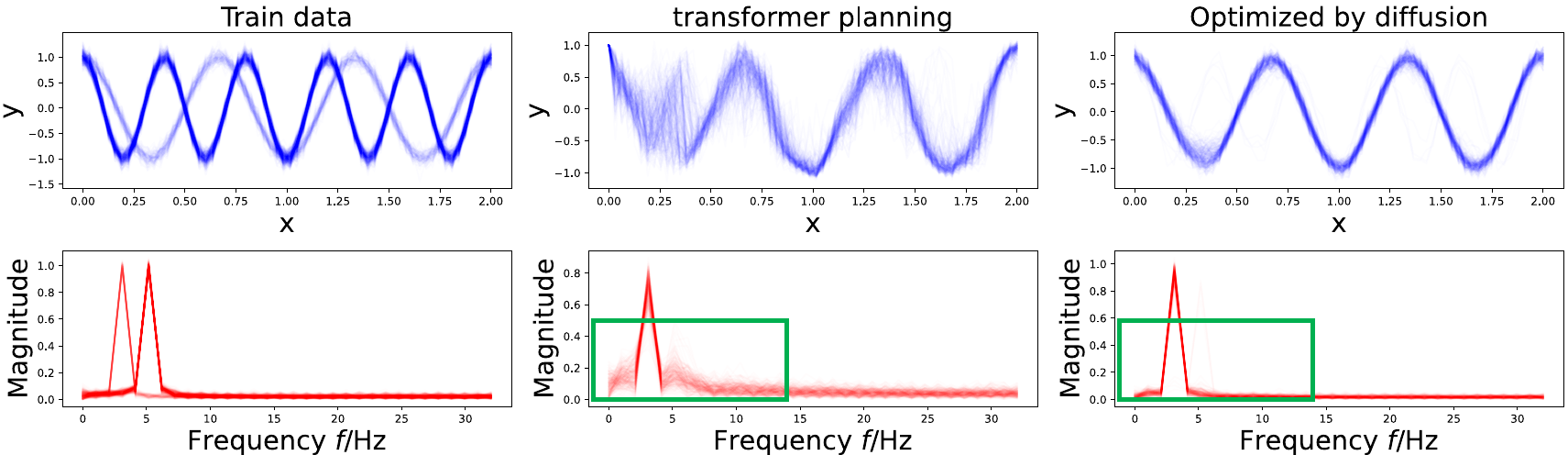}
    \caption{Toy example for 1D trajectory planning.  The result shows that the diffusion model optimizes the details without affecting the primary distribution.}
    \label{fig2}
\end{figure}

We first validate our assumption through a bandit case: The diffusion sampling process undergoes optimization in the final few steps~(Figure~\ref{fig1}(a)), which can be used as an optimizer to improve the results of other methods~(e.g., BC-CVAE)~(Figure~\ref{fig1}(b)). Based on these assumptions, we explain how our method achieves performance advantages from a frequency perspective~(Figure~\ref{fig2}). In a 1D trajectory planning task, the train data consists of mixed frequency trajectories, planning with the transformer involves interfering frequencies, but the primary frequency is correct. By incorporating a small number of diffusion steps for optimization, the interference frequencies are significantly reduced while maintaining the primary frequency. We can conclude that introducing the diffusion model as an optimizer does not change the main distribution of the trajectory but helps obtain the optimal solution through optimization. These two experiments demonstrate that our method simultaneously benefits from both efficiency and performance. The improvement in efficiency comes from a significant reduction in sampling steps, while the performance is enhanced by introducing the transformer for good initialization and combining it with the diffusion model optimization.


\subsection{Additional Experiment Results}

In this section, We present a comprehensive comparison with previous methods (\Cref{tab:result}). We provide additional experimental results (Table \ref{more result}), including the Maze2D tasks and the relocate, hammer and door tasks in Adroit. We present the training curves (\Cref{fig:enter-label}) for three representative tasks to demonstrate the stability and convergence of our approach. We conduct a more detailed comparison to analyze the impact of different optimization steps on the performance of Trajectory Diffuser. We provide a tuning range for the hyperparameters (scalar $w$, horizon $H$) during model training, and the results (\Cref{tab:w}, \Cref{tab:h}) demonstrate that our approach exhibits low sensitivity to hyperparameters. Furthermore, we showcase different performances under various returns-to-go (RTG) conditions (\Cref{tab:rtg}), thereby indicating the capability of our method to extend to other condition-based tasks.

\noindent\textbf{Comparison of methods with a wider range.} We present a comprehensive comparison with previous methods in Offline RL, including behavior cloning, model-based approaches, TD learning methods, and sequence modeling methods. The experimental results \Cref{tab:result} demonstrate that our approach is highly competitive overall and achieves state-of-the-art results in sequence modeling-based methods. Furthermore, our method shows remarkable improvements in inference speed.

\begin{table*}[!ht]
    \small
    \centering
    \caption{\textbf{The performance of different offline RL algorithms and Trajectory Diffuser.} The results are categorized based on method categorization, and the best performance within each category is highlighted.}
    \resizebox{\linewidth}{!}{
    \begin{tabular}{c||ccc||ccc||ccc||c||c}
    \toprule[1.2pt]
        \textbf{Dataset} & \multicolumn{3}{c||}{Med-Expert} & \multicolumn{3}{c||}{Medium} & \multicolumn{3}{c||}{Med-Replay}& \multirow{2}{*}{Average} & \multirow{2}{*}{Categorization} \\ \cmidrule{1-10}
        \textbf{Environment} & HalfCheetah & Hopper & Walker2d & HalfCheetah & Hopper & Walker2d & HalfCheetah & Hopper & Walker2d &  &  \\ \midrule
        BC & 55.2 & 52.5 & 107.5 & 42.6 & 52.9 & 75.3 & 36.6 & 18.1 & 26 & 51.9 & Behavior Cloning \\ \midrule
        MOReL & 53.3 & 108.7 & 95.6 & 42.1 & 95.4 & 77.8 & 40.2 & 93.6 & 49.8 & 72.9 & Model-based \\ \midrule
        BRAC & 41.9 & 0.9 & 81.6 & 46.3 & 31.3 & 81.1 & 47.7 & 0.6 & 0.9 & 36.9 & \multirow{4}{*}{Temporal Difference} \\ 
        CQL & 91.6 & 105.4 & 108.8 & 44 & 58.5 & 72.5 & 45.5 & 95.0 & 77.2 & 77.6 &  \\ 
        IQL & 86.7 & 91.5 & 109.6 & 47.4 & 66.3 & 78.3 & 44.2 & 94.7 & 73.9 & 77.0 &  \\ 
        DQL & 96.8 & 111.1 & 110.1 & 51.1 & 90.5 & 87.0 & 47.8 & 101.3 & 95.5 & \textbf{88.0} &  \\ \midrule
        DT & 86.8 & 107.6 & 108.1 & 42.6 & 67.6 & 74.0 & 36.6 & 82.7 & 66.6 & 74.7 & \multirow{5}{*}{Sequence Modeling} \\ 
        TT & 95.0 & 110.0 & 101.9 & 46.9 & 61.1 & 79.0 & 41.9 & 91.5 & 82.6 & 78.9 & ~ \\ 
        Diffuser & 88.9 & 103.3 & 106.9 & 42.8 & 74.3 & 79.6 & 37.3 & 93.6 & 70.6 & 77.5 & ~ \\ 
        DD & 90.6 & 111.8 & 108.8 & 49.1 & 79.3 & 82.5 & 39.3 & 100.0 & 75.0 & 81.8 & ~ \\ 
        AdaptDiffuser  & 89.6 & 111.6 & 108.2 & 44.2 & 96.6 & 84.4 & 38.3 & 92.2 & 84.7 & 83.4 & ~ \\ 
        TD~(Ours) & 92.2 $\pm$0.8 & 113.6 $\pm$1.5 & 109.3 $\pm$0.8 & 44.3 $\pm$1.1 & 97.1 $\pm$1.5 & 83.5 $\pm$2.2 & 41.3 $\pm$0.8 & 94.3 $\pm$3.2 & 82.5 $\pm$2.6 & \textbf{84.2} \\ 
        \bottomrule[1.2pt]
    \end{tabular}
    }
    \label{tab:result}
\end{table*}

\noindent\textbf{Hyperparameters tuning range.} We conducted ablation experiments to demonstrate the tuning range of two important hyperparameters: scalar $w$ and horizon $h$. The experimental results are presented in \Cref{tab:w} and \Cref{tab:h}. It is evident that different values of $w$ and $h$ do have some impact on the model's performance. However, we observed that the variations resulting from these impacts are not particularly pronounced or significant.

\noindent\textbf{Conditional guidance.} We evaluate whether the desired condition guides the performance of the policy. Specifically, we are conducting ablation experiments (\Cref{tab:rtg}) by varying the returns-to-go (RTG) parameter. The RTG parameter represents the best model performance, and we multiply it by different discount factors to signify the various goals the policy needs to achieve. The experimental results indicate that different conditions have different guiding effects on the policy, suggesting that our method can be extended to other condition-based tasks.

\begin{table}[b]
    \small
    \centering
     \caption{Ablation of horizon $w$, evaluation on Medium-Expert Dataset.}
    \resizebox{\linewidth}{!}{
    \begin{tabular}{lllllllll}
    \toprule[1.2pt]
        Env & $w = 1.0 $  & $w = 1.1$  & $w = 1.2$ & $w = 1.3 $ & $w = 1.4$ & $w = 1.5$ & $w = 2.0$ & $w = 3.0$ \\ \midrule
        Hopper & 109.1         $\pm$7.4 & 112.3          $\pm$2.5 & \textbf{113.2}      $\pm$0.6 & 112.4  $\pm$0.9 & 108.5  $\pm$2.7 & 106.7  $\pm$8.6 & 109.9  $\pm$7.0 & 112.2  $\pm$1.2 \\ 
        walker2d & 108.3       $\pm$2.7 & 108.4          $\pm$0.3 & \textbf{108.9}      $\pm$0.2 & 107.8  $\pm$0.6 & 107.9  $\pm$0.4 & 108.2  $\pm$0.3 & 108.0  $\pm$0.4 & 108.1  $\pm$0.5 \\ 
        halfcheetah & 91.1     $\pm$0.9 & \textbf{92.1}           $\pm$1.5 & 90.7       $\pm$1.4 & 90.7   $\pm$1.2 & 89.7   $\pm$0.7 & 91.2   $\pm$0.5 & 87.0   $\pm$7.2 & 90.0   $\pm$2.6 \\ \bottomrule[1.2pt]
    \end{tabular}
    }
    \label{tab:w}
\end{table}

\begin{figure}[ht]
    \centering
        \begin{minipage}{0.42\textwidth}
            \small
            \centering
            \captionof{table}{Ablation of horizon $H$, evaluation on Medium-Expert Dataset.}
            \resizebox{\textwidth}{!}{
            \begin{tabular}{lllll}
            \toprule[1.2pt]
                Env & $H$ = 8 & $H$ = 16 & $H$ = 24 & $H$ = 32 \\ \midrule
                Hopper & 105.4 $\pm$16.1 & \textbf{113.6} $\pm$2.3 & 112.2 $\pm$0.9 & 112.3 $\pm$0.7 \\ 
                Walker2d & 108.7 $\pm$0.4 & 108.3 $\pm$0.2 & 108.7 $\pm$0.3 & \textbf{109.1} $\pm$0.4 \\ 
                Halfcheetah & \textbf{92.3} $\pm$0.7 & 90.9 $\pm$1.5 & 90.0 $\pm$4.2 & 82.2 $\pm$20 \\ 
            \bottomrule[1.2pt]
            \end{tabular}
            }            
        \label{tab:h}
    \end{minipage}
        \hspace{0.05\textwidth}
    \begin{minipage}{0.5\textwidth}
        \small
        \centering
        \captionof{table}{Ablation of returns-to-go (RTG), evaluation on Medium-
Expert Dataset.}
        \resizebox{\textwidth}{!}{
        \begin{tabular}{llllll}
        \toprule[1.2pt]
            Env & 1 * RTG & 0.75 * RTG & 0.5 * RTG & 0.25 * RTG & 0 * RTG \\ \midrule
            Hopper & 113.6 $\pm$2.3 & 103.0 $\pm$20.7 & 84.9 $\pm$10.3 & 48.2 $\pm$16.1 & 13.2 $\pm$0.6 \\ 
            Walker2d & 108.7 $\pm$0.4 & 75.6 $\pm$3.5 & 76.4 $\pm$2.0 & 30.3 $\pm$25.4 & 27.9 $\pm$27.4 \\
            Halfcheetah & 91.8 $\pm$1.4 & 90.8 $\pm$0.9 & 42.7 $\pm$1.0 & 42.8 $\pm$1.2 & 42.1 $\pm$0.8 \\
        \bottomrule[1.2pt]
        \end{tabular}
        }
        
        \label{tab:rtg}
    \end{minipage}     
\end{figure}

\noindent\textbf{Comparison with more tasks.} We assess the performance of our method on more demanding tasks (\Cref{more result}), include Maze2D and Adroit. The Maze2D task tests the policy's ability to perceive over long horizons and make decisions based on specific conditions. In the Adroit task, the policy must handle out-of-distribution data and exhibit regularization capabilities to prevent observations from straying away from the desired region.
\begin{itemize}
    \item \textbf{Evaluation on Maze2D.} In this experiment, we evaluate the performance of our method (TD) and several baselines in a sparse reward setting. The sparse reward setting poses a challenge for the agent to discover successful trajectories, particularly in offline RL tasks where the agent cannot explore new experiences through direct interaction with the environment. The results indicate that our method is capable of effectively tackling tasks that require long-horizon planning due to a sparse reward structure.

    \item \textbf{Evaluation on Adroit.} Due to the predominantly human behavior-driven collection of Adroit datasets, the state-action region represented by the offline data is often limited in scope. Consequently, it is crucial to impose robust policy regularization to ensure that the agent remains within the desired region.
\end{itemize}

\begin{table}[ht]
    \centering
        \caption{Offline RL algorithm comparison on Maze2D dataset and Adroit dataset.}
    \resizebox{0.5\textwidth}{!}{
    \begin{tabular}{lllllll}
    \toprule[1.2pt]
        Dataset& Environment& BC & AWR & BCQ & CQL &TD(Ours)\\ \midrule
        umaze & maze2d & 3.8 & 1.0   &  12.9  &   5.7 & \textbf{118.9} \\ 
        medium & maze2d & 30.0 &  7.6 &  8.3  &   5.0 & \textbf{56.1} \\ 
        large & maze2d & 5.0 &  23.7  &  6.2 &    12.5& \textbf{39.9} \\ \midrule
        \multicolumn{2}{c}{\textbf{Average}} & 12.9 & 10.8 & 9.1 &  7.7 & \textbf{71.6} \\ 
        \midrule
        human & relocate &  0.0    & 0.0  & -0.1   & \textbf{0.2} & 0.0 \\ 
        expert & relocate & 101.3 & 91.5 &  41.6  & 95.0 & \textbf{111.1} \\ 
        cloned & relocate & -0.1  & -0.2 &  0.3  & -0.1 & \textbf{0.2} \\ \midrule
        human & hammar & 1.5 & 4.4 &1.2 &  0.5  & \textbf{13.4} \\ 
        expert & hammar & 125.6  &   39.0 & 107.2  & 86.7 & \textbf{127.6} \\ 
        cloned & hammar & 0.8   &   0.4  &  0.4 & 2.1 & \textbf{17.8} \\  \midrule
        human & door & 0.5      &   0.4  &  0.0 & 9.9 & \textbf{12.2} \\ 
        expert & door & 34.9    &   102.9 &  99.0  & 101.5 & \textbf{105.3} \\ 
        cloned & door & -0.1    &   0.0   & 0.0 & 0.4 & \textbf{10.6} \\ \midrule
        \multicolumn{2}{c}{\textbf{Average}} & 29.4  &26.1  & 27.7      & 33.3 & \textbf{44.2} \\
        \bottomrule[1.2pt]
    \end{tabular}    
    }
    \label{more result}
\end{table}

\noindent\textbf{Training stability.} We have provided training curves for three representative tasks, as depicted in Figure \ref{fig:enter-label}. The term \textit{Average} denotes the average performance, while \textit{Best} represents the best performance achieved.
For the hopper and walker2d tasks, the training process exhibits stability, with consistent performance improvements over time.
However, in the case of the halfcheetah task, the training stability is comparatively poorer. Despite this, the training process eventually converges to its maximum value, indicating that the model still achieves the desired performance level.

\begin{figure*}[!h]
    \centering
    \subfloat{\includegraphics[width=0.33\textwidth]{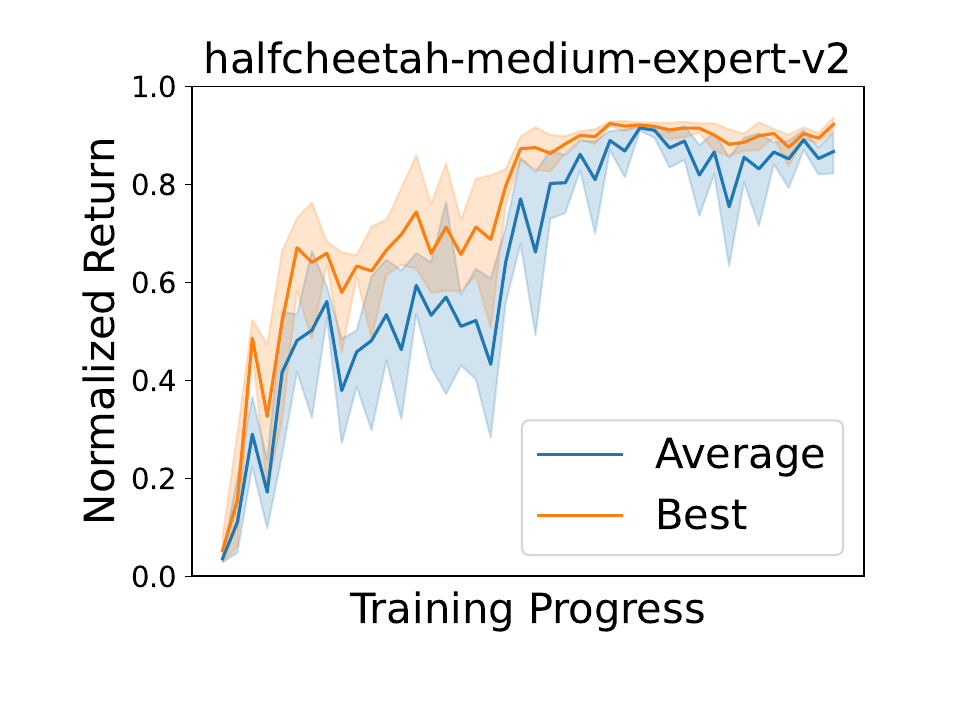}}
    \subfloat{\includegraphics[width=0.33\textwidth]{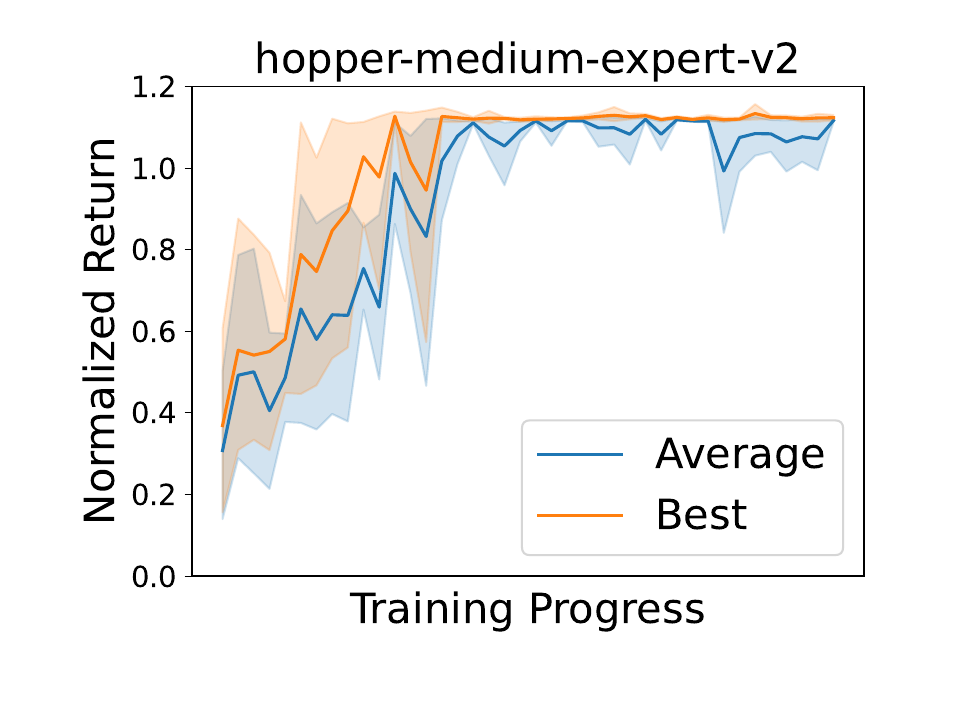}}
    \subfloat{\includegraphics[width=0.33\textwidth]{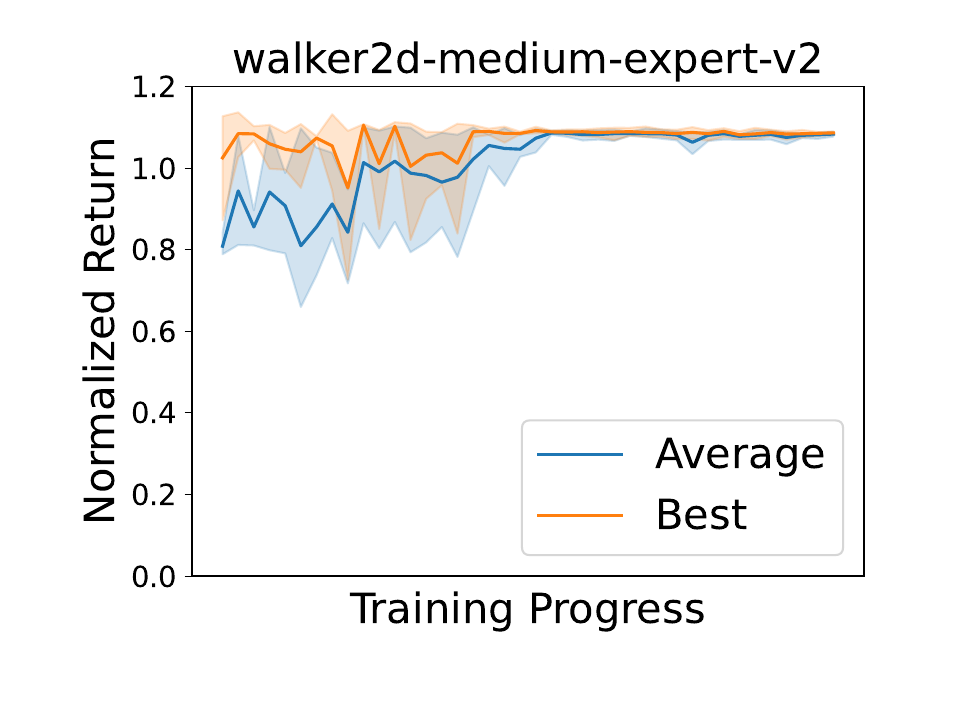}}
    \caption{Training curves for Trajectory Diffuser on three representative environments.}
    
    \label{fig:enter-label}
\end{figure*}

\subsection{Pseudocode of Trajectory Diffuser}
Pseudocode for trajectory planning with Trajectory Diffuser is shown in Algorithm \ref{fullalg}. 
\begin{algorithm*}[t]
\renewcommand{\algorithmicrequire}{\textbf{Input:}}
\renewcommand{\algorithmicensure}{\textbf{Requirements:}}
\renewcommand{\algorithmiccomment}[1]{\hfill $\triangleright$ #1}

\caption{Planning With Trajectory Diffuser(for continuous actions)}
\label{fullalg}
\begin{algorithmic}[1]
\REQUIRE history length $L$, imporve steps $K$, planning steps $C$, reward-to-go $rtg$, guidance scale $\omega$
\ENSURE sequence model $transformer$, Noise model $\epsilon_\theta$ and inverse dynamics $f_\phi$

\STATE Initialize $h\gets Queue(length=L), r\gets Queue(length=L), t\gets 0$ \COMMENT{Maintain a history of length C}
\WHILE{not done}
    \STATE Observe state $s_t; h.insert(s_t);$
    \STATE Returns-to-go $rtg; r.insert(rtg);$
    \STATE \textbf{// Planning stage}
    \STATE extract history states: $states \gets h$
    \STATE extract history rtgs: $rtgs \gets r$
    \FOR{$c = C \dots 1$}
        \STATE $next\_state, next\_rtg = transformer(states[-L:], rtgs[-L:])$
        \STATE append $next\_state$ to $states$
        \STATE append $next\_rtg$ to $rtgs$
    \ENDFOR
    \STATE \textbf{// Improve stage}
    \IF{improve stage} 

    \STATE Initialize $\boldsymbol{x}_K(\tau) \gets \text{AddNoise}(states[t:t+c])$
    \FOR{$k = K \dots 1$}
        \STATE $\boldsymbol{x}_k(\tau)[0] \gets states[t]$ \COMMENT{Constrain plan to be consistent with current state}
        \STATE $\hat{\epsilon}\gets \epsilon_\theta(\boldsymbol{x}_k(\tau), k) + \omega(\epsilon_\theta(\boldsymbol{x}_k(\tau), y, k)-\epsilon_\theta(\boldsymbol{x}_k(\tau), k))$ \COMMENT{Classifier-free guidance}
        \STATE $(\mu_{k-1}, \sum_{k-1})\gets \text{DeNoise}(\boldsymbol{x}_{k}(\tau), \hat{\epsilon})$
        \STATE $\boldsymbol{x}_{k-1}\sim \mathcal{N}(\mu_{k-1}, \alpha\sum_{k-1})$
    \ENDFOR
    \STATE $states[t:t+c] \gets \boldsymbol{x}_0(\tau)$
    \ENDIF
    \STATE Extract$(s_t, s_{t+1})$ from $states$
    \STATE Execute $a_t \gets f_\phi(s_t, s_{t+1})$; $t \gets t+1$
\ENDWHILE
\end{algorithmic}
\end{algorithm*}

To utilize the Trajectory Diffuser for planning purposes, an initialization step involves establishing two queues responsible for maintaining a record of historical states and rewards. Subsequently, a transformer model is employed to perceive the historical states, while trajectory planning is carried out through an autoregressive prediction technique. It is worth noting that the autoregressive approach is prone to error accumulation, leading to the emergence of suboptimal trajectories and out-of-distribution scenarios. In order to mitigate these challenges, a diffusion model is leveraged to capture the underlying distribution of trajectories, enabling their subsequent adjustment towards the optimal trajectory. Finally, an inverse dynamics model is applied to extract the actionable actions.

\subsection{Implementation Details}
In this section, we describe various architectural and hyperparameter details: 
\begin{itemize}
    \item For the transformer model, we employ $4$ transformer block layers and $4$ attention heads. The embedding dimension is set to $256$. Finally, we have utilized the Rectified Linear Unit (ReLU) activation function as the non-linear activation layer.
    \item We represent the noise model $\theta$ with a temporal U-Net \cite{janner2022planning}, consisting of a U-Net structure with 6 repeated residual blocks. Each block consisted of two temporal convolutions, each followed by a group norm \cite{wu2018group}, and a final Mish nonlinearity \cite{mish1908self}. Timestep and condition embeddings, both 64-dimensional vectors, are produced by separate 2-layered MLP (with 128 hidden units and Mish nonlinearity) and are concatenated together before getting added to the activations of the first temporal convolution within each block. 
    \begin{table}[ht]
    \centering
    \caption{Hyperparameter settings of all selected tasks.}
    \begin{tabular}{llccccc}
    \toprule[1.2pt]
        Dataset & Environment & learning rate & horizon $H$ & improve step $k$ &  scalar $w$ & returns-to-go $\hat{R}$ \\ \midrule
        Med-Expert & HalfCheetah & $2\times 10^{-4}$ & 8 & 5 & 1.1 & 11000 \\ 
        Medium & HalfCheetah & $2\times 10^{-4}$ & 8 & 5 & 1.1 & 5300 \\ 
        Med-Replay & HalfCheetah & $2\times 10^{-4}$ & 8 & 5 & 1.1 & 5300\\ \midrule
        Med-Expert & Hopper & $2\times 10^{-4}$ & 16 & 5 & 1.2 & 3700\\ 
        Medium & Hopper & $2\times 10^{-4}$ & 16 & 5 & 1.2 & 3100\\ 
        Med-Replay & Hopper & $2\times 10^{-4}$ & 16 & 5 & 1.2 & 3100\\ \midrule
        Med-Expert & Walker2d & $2\times 10^{-4}$ & 32 & 5 & 1.2 & 5100\\ 
        Medium & Walker2d & $2\times 10^{-4}$ & 32 & 5 & 1.2 & 4200\\ 
        Med-Replay & Walker2d & $2\times 10^{-4}$ & 32 & 5 & 1.2 & 4200\\ \midrule
        human  & pen & $1\times 10^{-4}$ & 16 & 2 & 1.3 & 6000\\ 
        expert  & pen & $1\times 10^{-4}$ & 16 & 2 & 1.3 & 6000\\ 
        cloned  & pen & $1\times 10^{-4}$ & 16 & 2 & 1.3 & 6000\\ \midrule
        Partial & Kitchen & $1\times 10^{-4}$ & 32 & 5 & 1.2 & 500\\ 
        Mixed & Kitchen & $1\times 10^{-4}$ & 32 & 5 & 1.2 & 400\\
        \bottomrule[1.2pt]
    \end{tabular}
    
    \label{Hyper}
\end{table}
    \item We represent the inverse dynamics $f_\phi$ with a 2-layered MLP with 256 hidden units and ReLU activations.
    \item We train $\theta$ and $f_\phi$ using the Adam optimizer \cite{kinga2015method} and batch size of 32 for $2e6$ train steps. 
    \item We choose the probability $p$ of removing the conditioning information to be $0.5$. 
    \item We use $K = 100$ diffusion steps for training.
    \item We use discount factor of $1.0$
    \item The details hyperparameter is shown in Table \ref{Hyper}
\end{itemize}

\subsection{More Discussion about Limitations}
We summarize some potential limitations of Trajectory Diffuser:
\begin{itemize}
    \item It circumvents exploration by focusing on offline sequential decision-making, but online fine-tuning of the Trajectory Diffuser could be incorporated to enable exploration using state-sequence model entropy.
    \item The model's applicability is currently limited to state-based environments, but extending it to image-based environments by performing diffusion in the latent space could enhance its capabilities.
    \item The diffusion models, including the Trajectory Diffuser, are prone to overfitting in scenarios with limited data, potentially limiting their effectiveness in certain applications.
    \item While we proposed using the diffusion model as an optimizer for trajectory regulation, we did not redesign the structure of the diffusion model.

\end{itemize}

\end{document}